\RequirePackage{fix-cm}
\documentclass[smallextended]{svjour3} 

\usepackage{graphicx}
\usepackage[export]{adjustbox}
\usepackage{array}
\usepackage{arydshln}
\usepackage{tabularx}
\usepackage{multirow}
\usepackage{amsmath}
\usepackage{mathtools}
\usepackage{color}
\usepackage{url}

\graphicspath{{./images/}}

\journalname{Multimedia Tools and Applications}

\DeclarePairedDelimiter{\floor}{\lfloor}{\rfloor}

\begin{document}

\title{Automatic Separation of Compound Figures\break in Scientific Articles}
\titlerunning{Automatic Separation of Compound Figures}

\author{Mario Taschwer \and
	Oge Marques}

\institute{Mario Taschwer \at
              ITEC, Klagenfurt University (AAU), Austria \\
              \email{mario.taschwer@aau.at}           
           \and					
          Oge Marques \at
              Florida Atlantic University (FAU), Boca Raton, FL, USA \\
              \email{omarques@fau.edu}                     
}


\maketitle

\section*{Abstract}
Content-based analysis and retrieval of digital images found in scientific articles is often hindered by images consisting of multiple subfigures (compound figures). We address this problem by proposing a method to automatically classify and separate compound figures, which consists of two main steps: (i) a supervised compound figure classifier (CFC) discriminates between compound and non-compound figures using task-specific image features; and (ii) an image processing algorithm is applied to predicted compound images to perform compound figure separation (CFS). Our CFC approach is shown to achieve state-of-the-art classification performance on a published dataset. Our CFS algorithm shows superior separation accuracy on two different datasets compared to other known automatic approaches. Finally, we propose a method to evaluate the effectiveness of the CFC-CFS process chain and use it to optimize the misclassification loss of CFC for maximal effectiveness in the process chain.

\keywords{multipanel figure separation \and document image understanding}

\section{Introduction}

\begin{figure}
\begin{tabular}{cc}
\includegraphics[width=0.37\textwidth]{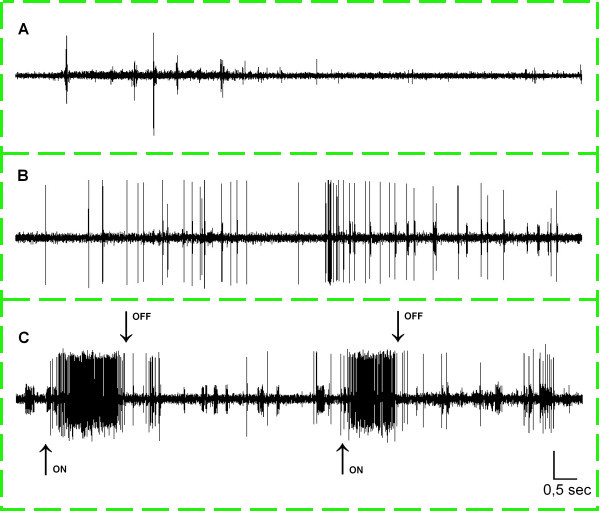}
&
\includegraphics[width=0.53\textwidth]{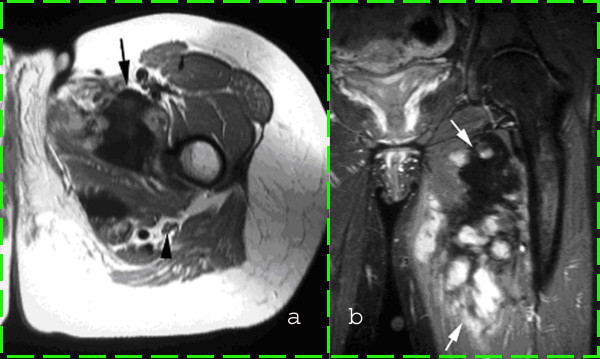}\\
(a) & (b)
\end{tabular}
\caption{Sample compound images (of the ImageCLEF 2015 CFS dataset \cite{Herrera2015}) suitable for two different separator detection algorithms. Subfigures are separated by (a) white{\-}space, (b) a vertical edge. Dashed lines represent the expected output of CFS.}
\label{fig:sample-compound-images}
\end{figure}

The work described in this paper is motivated by the realization that articles in scientific publications contain a substantial amount of figures consisting of two or more subfigures, which could be treated as separate images for the purpose of automatic content-based analysis or indexing for retrieval. Figure~\ref{fig:sample-compound-images} shows two examples of such \emph{compound figures} found in a dataset of article images of the biomedical literature. Based on published datasets drawn from open access biomedical literature, it has been estimated that between 40\% and 60\% of figures occurring in articles are compound figures \cite{Apostolova2013,Chhatkuli2013,Herrera2013}.

In this paper we address the problem of automatically recognizing and separating compound figures in a collection of article images by breaking it into two subproblems: \emph{compound figure classification} (CFC) and \emph{compound figure separation} (CFS). CFC is a binary classification problem that aims at discriminating between compound and non-compound figures given an article image. CFS is the problem of determining the bounding boxes of all subfigures of a given compound figure. Algorithms solving the CFC and CFS problems are naturally combined into a \emph{CFC-CFS process chain} that receives arbitrary article images as input and delivers bounding boxes of subfigures (or of single figures) at the output. Images classified as \emph{compound} by the CFC algorithm are further processed by CFS, whereas for images predicted as \emph{non-compound} by the CFC a bounding box covering the entire image is produced.

For CFC, we propose several global image features designed to capture the existence of edges or whitespace that could potentially separate subfigures and use them with well-known supervised machine learning algorithms. For CFS, we designed an image processing algorithm comprising distinct modules for detecting two types of separators between subfigures: (1) homogeneous rectangular areas of whitespace spanning the entire image width or height, which we call \emph{separator bands} (shown in Fig.~\ref{fig:sample-compound-images}(a)); and (2) \emph{separator edges} spanning the entire image width or height, which may arise from borders drawn around subfigures or from adjacent subfigures ``stitched together'' as shown in Fig.~\ref{fig:sample-compound-images}(b). The proposed CFS algorithm internally uses a separate binary classifier (independent from CFC) to decide which of the two separator detection modules to apply to a given compound image. Based on the observation that compound images containing graphical illustrations (such as diagrams and charts) often contain separator bands, whereas most subfigures in other compound images show rectangular border edges, we train the internal CFS classifier to discriminate between graphical illustrations and other article images and call it \emph{illustration classifier}.

This paper is based on previous work \cite{Taschwer2015,Taschwer2016} and provides the following additional research contributions: 

\begin{enumerate}
  \item 
  It proposes novel image features for compound figure classification, which can be efficiently extracted and achieve state-of-the-art CFC performance using well-known classifier algorithms.
  
  \item 
  It demonstrates that the proposed CFS algorithm outperforms state-of-the-art automatic and semi-automatic CFS approaches on two recently published biomedical datasets.

  \item 
  It establishes a method to evaluate CFC-CFS chain effectiveness, which is applied to investigate the effect of various CFC implementations in the chain.
\end{enumerate}


\subsection{Motivation and Context}

From a problem-oriented point of view, research on automatic compound figure separation is motivated by the fact that compound figures hamper content-based analysis and indexing of article images for retrieval, because global image features extracted from a compound image are a mixture (often an average) of the same features extracted from the subfigures only, leading to reduced discriminative power of these features on compound images. The situation may be slightly better for local image features, which capture the existence of certain texture or shape patterns in small image regions, but the predominant way of aggregating local features of an image in a Bag of Visual Words representation \cite{Sivic2003} still suffers from the additive effect of including local features from all subfigures. Moreover, subfigures of a given compound image usually convey different semantic information that may be relevant for retrieval, although the compound figure establishes a common semantic context for subfigures.

From a historical perspective, the research community showed little interest in the CFC and CFS problems until ImageCLEF 2013, where a CFS task was introduced as one of the challenges in the biomedical domain \cite{Herrera2013}. Task organizers provided training and test datasets, and evaluated CFS results submitted by participants for the test dataset. It is presumably this provisioning of datasets and of a CFS evaluation method that stimulated research on CFC and CFS problems in recent years. Since we are not aware of any CFC and CFS datasets available to the research community other than the ones described in Section~\ref{sec:datasets}, our experiments are limited to the biomedical domain, although our proposed algorithms should be applicable to other scientific domains as well.

\subsection{Related Work}
\label{sec:related-work}

There is little work about the CFC problem in the literature, but research interest may grow due to a CFC task introduced recently at ImageCLEF 2015 \cite{Herrera2015}. From the two participating groups of this task, Pelka and Friedrich \cite{Pelka2015} achieved best results (an accuracy of 85.4\%) using both textual and visual features with a random forest classifier. Textual features were extracted from image captions, and visual features were derived from detected separator bands and a bag-of-keypoints representation of dense-sampled SIFT keypoints. Their submitted variant using visual features only resulted in 72.5\% accuracy. Wang et al. \cite{Wang2015}, the other participating group in the ImageCLEF 2015 CFC task, achieved 82.8\% accuracy using visual features only. They used an unsupervised approach consisting of connected component analysis followed by separator band detection. A separate evaluation run using connected component analysis only resulted in 82.5\% accuracy, so separator band detection had an almost negligible effect in the combined approach.

Prior to ImageCLEF 2015, Yuan and Ang \cite{Yuan2014} proposed a 3-class classifier discriminating between photographs, non-photographs, and compound images containing both photographs and non-photographs. The classifier was used in a process chain for CFS, but it is not clear whether the classifier output should affect CFS operation. Experiments did not include classifier evaluation.

An established research field whose techniques could potentially be useful for CFC is document image classification \cite{Chen2007}. Although it deals with digital images of entire document pages (containing mostly text), some of the proposed methods -- such as block segmentation and physical layout analysis -- may also help CFC. However, there is no evidence of such utilization in the literature yet.

The success of deep learning techniques \cite{Bengio2009} or other advanced methods of representation learning \cite{Bengio2013} in image classification tasks during the last decade \cite{Krizhevsky2012,Srivastava2014}  suggest that they could also be applied to the CFC problem. However, we believe that available CFC training sets are still too small to obtain effective classifiers from deep learning methods, and we hope that recently proposed ``simple'' CFC methods (including ours) will help build larger training sets for advanced machine learning techniques.

Regarding the CFS problem, most existing approaches focus on the detection of separator bands \cite{Apostolova2013,Chhatkuli2013,Kitanovski2013,Yuan2014} and hence fail for compound images where subimages are stitched together without separator bands (see Fig.~\ref{fig:sample-compound-images}(b)). Apostolova et al. \cite{Apostolova2013} propose to solve the CFS problem using not only visual information contained in article images, but also textual information contained in images (extracted using OCR techniques) and in image captions. Since our proposed CFS algorithm does not use textual information, we compare it to their visual CFS algorithm (described as \emph{image panel segmentation}), which is part of a five-stage process chain and includes image markup removal. Image markup consists of text labels embedded in compound images that may be located in separator bands, exacerbating separator detection. Their CFS method~\cite{Apostolova2013} has recently been used in an approach to document image classification and retrieval by Simpson et al. \cite{Simpson2015}.

Chhatkuli et al. \cite{Chhatkuli2013} employ several image preprocessing techniques -- including binarization, border cropping, and image markup removal -- prior to detecting separator bands. Separator band detection is done recursively in horizontal direction first, followed by recursive detection of vertical separators, which may lead to limited separation of irregular subfigure structures. Separator candidates are filtered by a complex rule-based analysis step. Evaluation is performed using a self-constructed dataset and evaluation method based on separator locations, making a comparison with our approach infeasible.

Kitanovski et al. \cite{Kitanovski2013} participated in the ImageCLEF 2013 CFS task using an apparently simple approach based on separator band detection. They do not provide details and achieved an accuracy of 69\%.

Yuan and Ang \cite{Yuan2014} build upon the approach of Murphy et al. \cite{Murphy2001} and of Qian and Murphy \cite{Qian2008} and use a sliding window to compute intensity histograms of horizontal and vertical bands to detect (white) separator bands. Additionally, they used an edge-based approach involving Hough transform to separate overlayed zoom-in views from background figures, a case that is not considered in this work. They evaluated their CFS method on two self-constructed small datasets of about 180~compound figures each, but did not provide enough details to make their evaluation method reproducible.

A different approach to CFS is based on connected component analysis of binarized images, which,  however, is susceptible to over-segmentation, in particular for subfigures containing diagrams or charts. Shatkay et al. \cite{Shatkay2006} used such a technique for CFS in the context of document classification, but did not evaluate CFS effectiveness separately. The CFC approach proposed by Wang et al. \cite{Wang2015} determines subfigures using connected component analysis that could probably also be used in a CFS algorithm.

The approach of NLM (U.S. National Library of Medicine) \cite{Santosh2015} and our previous approach \cite{Taschwer2015}, both submitted to the ImageCLEF 2015 \cite{Herrera2015} CFS task, independently proposed to address compound images without separator bands by processing edge detection results. Besides algorithmic differences in edge-based separator detection, our approach incorporates a classifier to automatically select edge- or band-based separator detection, whereas NLM's approach uses manual image classification for evaluation.%
\footnote{We therefore call NLM's approach \cite{Santosh2015} \emph{semi-automatic}, although an automatic classifier could be easily integrated.}

\subsection{Structure of the Paper}

The proposed methods to address the CFC and CFS problems are described in Section~\ref{sec:approach}, including a way to improve the effectiveness of the CFC-CFS process chain (Section~\ref{sec:cfc-cfs-chain}). Section~\ref{sec:experiments} explains the experimental setup to evaluate our approach and, in particular, describes the datasets (Section~\ref{sec:datasets}) and evaluation methods (Section~\ref{sec:evaluation-methods}) used. Evaluation results are presented and discussed in the same section, separately for CFC (Section~\ref{sec:cfc-experiments}), CFS (Section~\ref{sec:cfs-experiments}), and the CFC-CFS process chain (Section~\ref{sec:cfc-cfs-experiments}). Section~\ref{sec:conclusion} concludes the paper and makes some suggestions for future work.

\section{Methods}
\label{sec:approach}

In the following two subsections, we describe our proposed approach to address the CFC and CFS problems, respectively. A technique to improve the effectiveness of the CFC-CFS process chain, given an imperfect CFC implementation, will be described in Section~\ref{sec:cfc-cfs-chain}.

\subsection{Compound Figure Classifier}
\label{sec:compound-fig-classifier}

Recognizing compound figures in a dataset of article images can be viewed as a binary classification problem. We address this problem by using hand-crafted image features and classical machine learning algorithms, because we consider the available training datasets as being too small for deep learning techniques (see Section~\ref{sec:related-work}), and we expect that the effect of limited classification accuracy on the CFC-CFS process chain can be partly compensated by biasing the classifier towards the \emph{compound} class (see Section~\ref{sec:cfc-cfs-chain}).

For CFC, we propose to use three types of image features determined separately for vertical and horizontal directions of a gray-scale image whose pixel values have been normalized to the range $[0,1]$. Each feature type is computed by aggregating each pixel line in direction $D$ (vertical or horizontal) to a single real number, resulting in a single \emph{projection vector} representing the image along direction $D'$ orthogonal to~$D$. The spatial distribution of values in the projection vector is then captured by a \emph{spatial profile vector} of fixed length. The final feature vector is formed by concatenating the horizontal profile vectors of the three feature types, followed by the corresponding vertical profile vectors.

The three feature types differ in how the projection vector is calculated: (1) \emph{mean} gray values along pixel lines, (2) \emph{variance} of gray values along pixel lines, and (3) one-dimensional \emph{Hough transform}, which counts the number of edge points aligned in direction~$D$ in a binary edge map of the input image. The binary edge map is produced by applying a gradient threshold on edges in direction~$D$ detected by the Sobel operator. Hough transform values are then normalized to the range $[0,1]$ using the image dimension in direction $D$ (width or height). Some of the spatial profile methods applied afterwards require \emph{quantization} of projection vectors, which is performed differently for the three feature types, using quantization parameters (positive integers) $p$, $q$, and $h$:

\begin{itemize}
\item Mean projection values are quantized into $p$ bins dividing $[0,1]$ into $p$ subintervals with lower bounds $1-2^{i-p}$ for $i = 1, 2, \dots, p$. The logarithmic scale for quantization should help to discriminate between high values (white separator bands) and others.
\item Variance projection values are quantized into $q$ bins dividing $[0,1]$ into $q$ subintervals with upper bounds $2^{i-q}$ for $i = 1, 2, \dots, q$. The logarithmic scale for quantization should help to discriminate between low-variance pixel lines (subfigure separators) and others.
\item Normalized Hough transform values are quantized into $h$ bins dividing $[0,1]$ into $h$ subintervals with lower bounds $1-2^{i-h}$ for $i = 1, 2, \dots, h$. The logarithmic scale for quantization should help to discriminate between Hough peaks (subfigure separators) and others.
\end{itemize}

We consider six spatial profile methods to produce profile vectors from projection vectors. Five of them require quantization of projection vectors and divide the vector of dimensionality~$N$ into $k$ \emph{spatial bins} of $\floor{N/k}$ or $\floor{N/k}+1$ adjacent positions. An additional profile method tries to capture the spatial structure of the projection vector using its Fast Fourier Transform (FFT).

\begin{itemize}
\item\emph{Profile 1:} A spatial bin is represented by the full normalized histogram of quantized projection values, resulting in $p$, $q$, or $h$ values per spatial bin.
\item\emph{Profile 2:} A spatial bin is represented by the quantized projection value that occurs most often (the mode). This value is then normalized to the range $[0,1]$.
\item\emph{Profile 3:} A spatial bin is represented by the relative frequency of the largest quantized projection value, resulting in a single number in the range $[0,1]$.
\item\emph{Profile 4:} A spatial bin is represented by its maximum quantized projection value, normalized to the range $[0,1]$.
\item\emph{Profile 5:} A spatial bin is represented by its average quantized projection value, normalized to the range $[0,1]$.
\item\emph{Profile 6:} The absolute values of the first $k$ low-frequency FFT coefficients of the projection vector are normalized by $1/N$, such that resulting values are constrained to the range $[0,1]$.
\end{itemize}

The dimensionality of feature vectors depends on parameters $k$, $p$, $q$, $h$, and on the profile method used for each of the three feature types, as presented in Table~\ref{tab:cfc-features-sets}. We denote a certain \emph{feature set} by three numbers $xyz$ representing the spatial profile numbers of mean projection ($x$), variance projection ($y$), and Hough Transform ($z$). A value of zero (e.g. $x=0$) means that the corresponding component of the feature vector has been dropped. For example, the feature set $034$ denotes a feature vector formed by concatenation of horizontal profiles of variance projection and Hough Transform, followed by corresponding vertical profiles. Both profile methods (3 and 4) represent a spatial bin by a single number, resulting in $k$ numbers per profile vector, $2k$ numbers for both horizontal profiles, and $4k$ numbers for the final feature vector.

\begin{table}
\caption{Dimensionality of various feature sets used for compound figure classification. $k$ denotes the number of spatial bins used to compute profile vectors. $p$, $q$, and $h$ are quantization parameters. The right-most column gives the dimensionality for parameter settings $k=16$, $p=5$, $q=8$, $h=3$.}
\label{tab:cfc-features-sets}
\centering
\begin{tabular}{lcr}
\hline
\textbf{Feature Set} & \textbf{Dimensionality} & \textbf{Example}\\
\hline
111 & $2*k*(p+q+h)$ & 512 \\
222 & $6*k$ & 96 \\
333 & $6*k$ & 96 \\
444 & $6*k$ & 96 \\
555 & $6*k$ & 96 \\
666 & $6*k$ & 96 \\
011 & $2*k*(q+h)$ & 352 \\
034 & $4*k$ & 64 \\
134 & $2*k*(p+2)$ & 224 \\
434 & $6*k$ & 96 \\
\hline
\end{tabular}
\end{table}

As classifier algorithms we use logistic regression, a linear support vector machine (SVM), and a non-linear SVM with a radial basis function kernel. 

\subsection{Compound Figure Separation}
\label{sec:cfs-method}

\begin{figure}
\includegraphics[width=\textwidth]{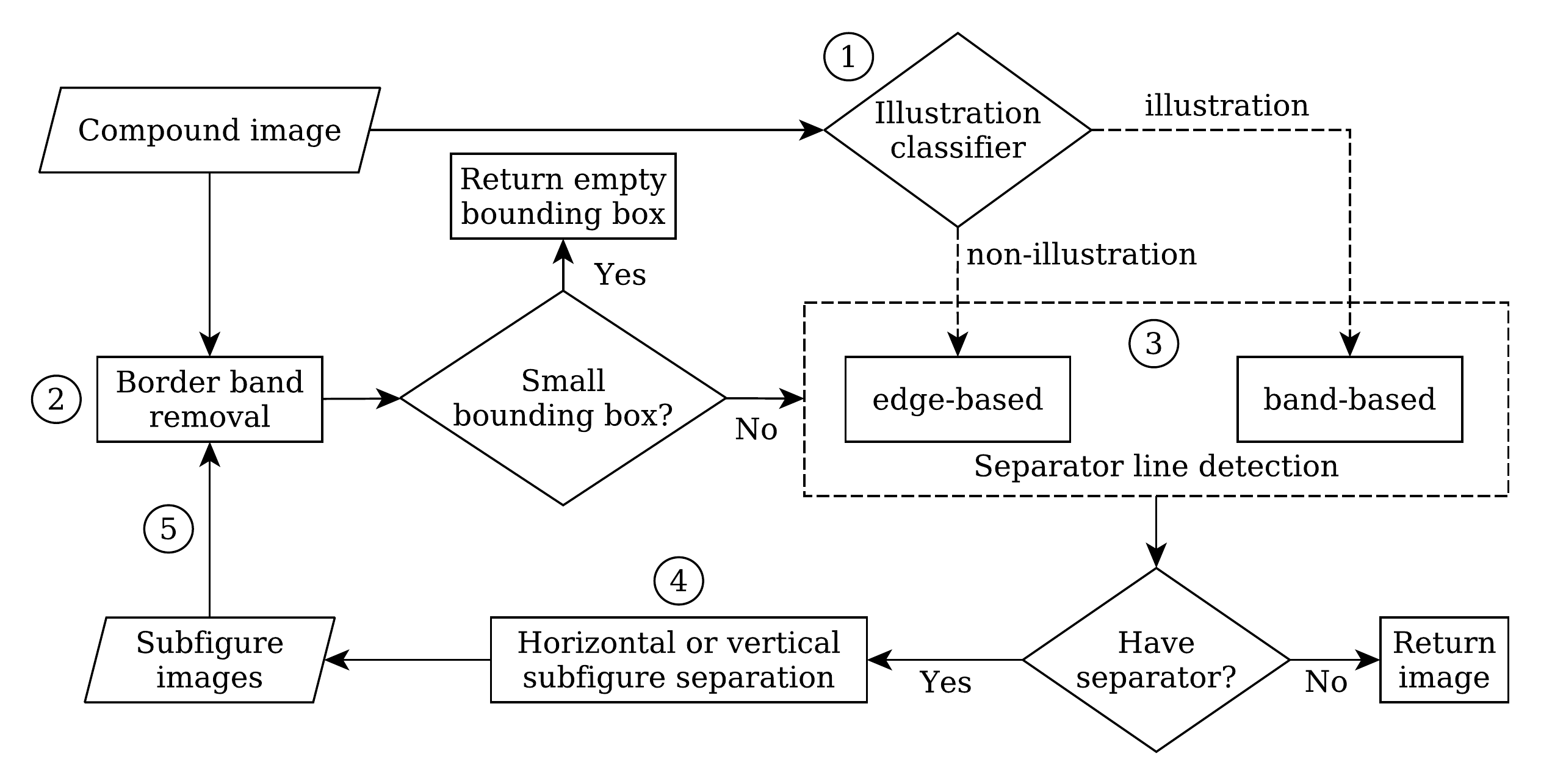}
\caption{Recursive algorithm for compound figure separation. Numbers denote the main algorithmic steps described in the beginning of Section~\ref{sec:cfs-method}.}
\label{fig:recursive-algorithm}
\end{figure}

Our approach to compound figure separation is a recursive algorithm (see Fig.~\ref{fig:recursive-algorithm}) which consists of the following steps: (1) classification of the compound image as illustration or non-illustration image, (2) removal of border bands, (3) detection of separator lines, (4) vertical or horizontal separation, and (5) recursive application to each subfigure image. The \emph{illustration classifier} is used to decide which of two separator line detection modules to apply: if the compound image is classified as an illustration image, the \emph{band-based} algorithm is applied, which aims at detecting separator bands between subfigures. Otherwise, the image is processed by the \emph{edge-based} separator detection algorithm, which applies edge detection and Hough transform to locate candidate separator edges. The algorithm selection is based on the assumption that edge-based separator detection is better suited for non-illustration compound images due to visible vertical or horizontal edges separating subfigures. Note that this assumption is not violated by non-illustration compound images with separator bands where subfigures have a visible rectangular border. The following four sections describe the illustration classifier, the main recursive algorithm, and the two separator detection modules in more detail.

\subsubsection{Illustration Classifier}
\label{sec:illustration-classifier}

The illustration classifier is used to decide which separator detection algorithm to apply to a given compound image. If the image is predicted to be a graphical illustration with probability greater than \texttt{decision\_threshold}, the band-based separator detection is applied, otherwise the edge-based separator module is used. This decision is made only once for each compound image, so all recursive invocations use the same separator detection algorithm.

Due to promising effectiveness for CFS in early experiments, we use four sets of global image features as classifier input, computed after gray-level conversion: (1) \emph{simple2} is a two-dimensional feature consisting of image entropy, estimated using a 256-bin histogram, and mean intensity; (2) \emph{simple11} extends \emph{simple2} by 9 quantiles of the intensity distribution; (3) \emph{CEDD} is the well-known color and edge directivity descriptor \cite{Chatzichristofis2008} (144-dimensional); and (4) \emph{CEDD\_simple11} is the concatenation of \emph{CEDD} and \emph{simple11} features (155-dimensional). 

As machine learning algorithms we consider support vector machines (SVM) with radial basis function kernel (RBF) and logistic regression. Although logistic regression is generally inferior to kernel SVM due to its linear decision boundary, it has the advantage of providing prediction probabilities, which allow us to tune the selection of separator detection algorithms using the \texttt{decision\_threshold} parameter.

\subsubsection{Recursive Algorithm}
\label{sec:recursive-algorithm}

Before applying the main algorithm (Fig.~\ref{fig:recursive-algorithm}) to a given compound figure image, it is converted to 8-bit gray-scale. \emph{Border band removal} detects a rectangular bounding box surrounded by a maximal homogeneous image region adjacent to image borders (border band). If the resulting bounding box is empty or smaller than \texttt{elim\_area} or if maximal recursion depth has been reached, an empty bounding box is returned, terminating recursion. The \emph{separator line detection} modules are invoked separately for vertical and horizontal directions, so they deal with a single direction~$\theta$ and return a list of corresponding separator lines. An empty list is returned if the respective image dimension (width or height) is smaller than \texttt{mindim} or if no separator lines are found. If the returned lists for both directions are empty, recursion is terminated and the current image (without border bands) is returned. The \emph{decision about vertical or horizontal separation} is trivial if one of both lists of separator lines is empty. Otherwise the decision is made based on the regularity of separator distances: locations of separator lines and borders are normalized to the range [0,1], and the direction (vertical or horizontal) yielding the lower variance of adjacent distances is chosen. Finally, the current figure image is divided into subimages along the chosen separation lines, and the algorithm is applied recursively to each subimage.

\subsubsection{Edge-based Separator Detection}
\label{sec:edge-based-algorithm}

\begin{figure}
\includegraphics[width=\textwidth]{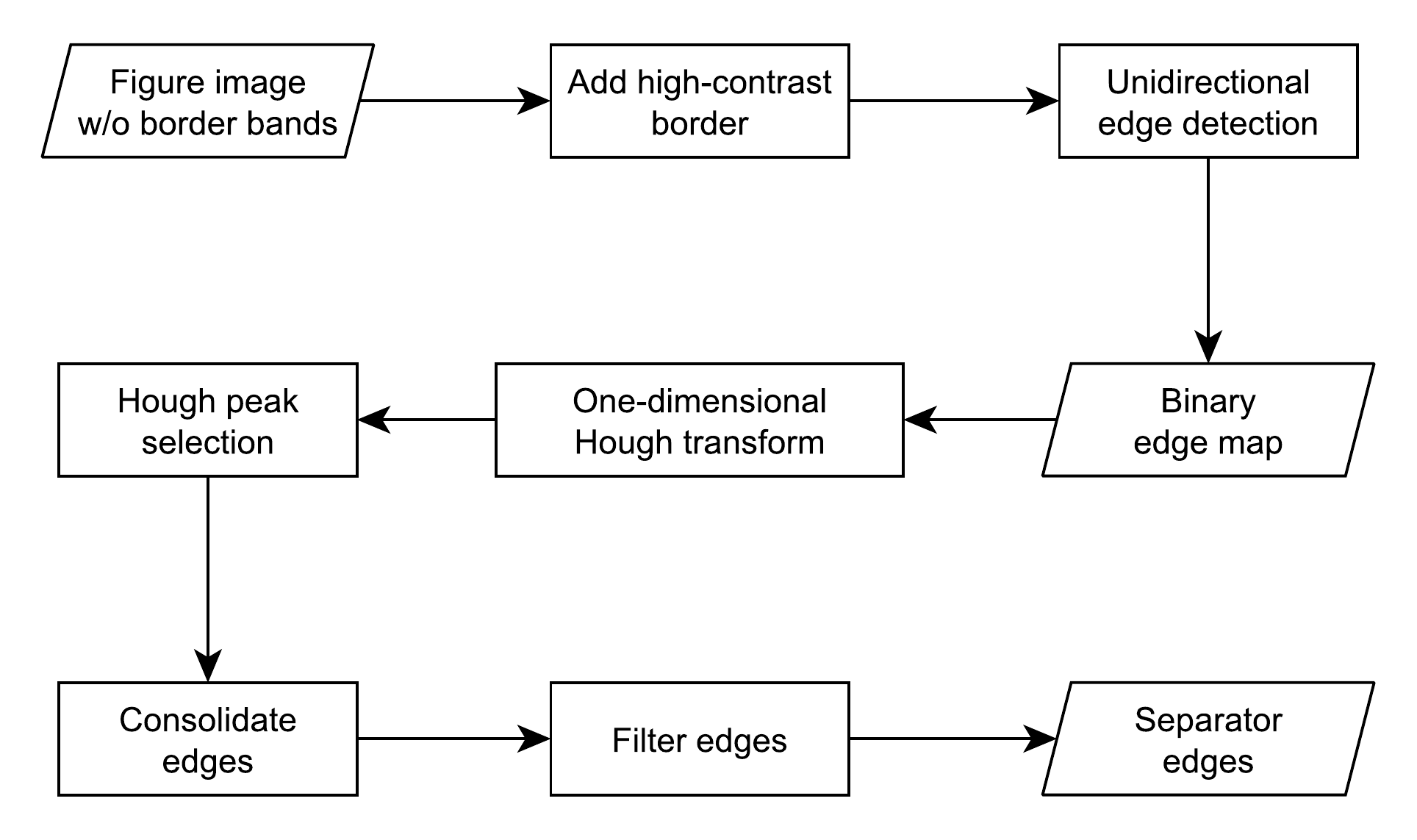}
\caption{Edge-based separator line detection.}
\label{fig:edge-based-algorithm}
\end{figure}

The edge-based separator line detection algorithm aims at detecting full-length edges of a certain direction $\theta$ (vertical or horizontal) in a given gray-scale image. It comprises the following processing steps depicted in Fig.~\ref{fig:edge-based-algorithm}: (1) unidirectional edge detection, (2) peak selection in one-dimensional Hough transform, and (3) consolidation and filtering of candidate edges.

Edge detection is implemented by a one-dimensional Sobel filter and subsequent thresholding (\texttt{edge\_sobelthresh}) to produce a binary edge map. The one-dimensional Hough transform counts the number of edge points aligned on each line in direction~$\theta$. So the peaks correspond to the longest edges, and their locations identify candidate separator edges. To make borders appear as strong Hough peaks, we add an artificial high-contrast border to the image prior to edge detection. Peaks are identified by an adaptive threshold $t$ that depends on the recursion depth~$k$ (zero-based), the maximal value~$m$ of the current Hough transform, and the fill ratio~$f$ of the binary edge map (fraction of non-zero pixels, $0\le f\le 1$), see (Eq.~\ref{eq:peak-threshold}). $\alpha$ and $\beta$ are internal parameters (\texttt{edge\_houghratio\_min} and \texttt{edge\_houghratio\_base}).

\begin{equation}\label{eq:peak-threshold}
h = \alpha * \beta^k\enspace ,\qquad
t = m * \left(h + (1-h)*\sqrt{f}\right)\enspace .
\end{equation}

\noindent The rationale behind these formulas is to cope with noise in the Hough transform. Hough peaks were observed to become less pronounced as image size decreases (implied by increasing recursion depth) and as the fill ratio~$f$ increases (more edge points increase the probability that they are aligned by chance). Equation~(\ref{eq:peak-threshold}) ensures a higher threshold in these cases. Additionally, as recursion depth increases, the algorithm should detect only more pronounced separator edges, because further figure subdivisions become less likely.

Hough peak selection also includes a similar regularity criterion as used for deciding about vertical or horizontal separation (see Section~\ref{sec:recursive-algorithm}): the list of candidate peaks is sorted by their Hough values in descending order, and candidates are removed from the end of the list until the variance of normalized edge distances of remaining candidates falls below a threshold (\texttt{edge\_maxdistvar}). Candidate edges resulting from Hough peak selection are then consolidated by filling small gaps (of maximal length given by \texttt{edge\_gapratio}) between edge line segments (of minimal length given by \texttt{edge\_lenratio}). Finally, edges that are too short in comparison to image height or width (threshold \texttt{edge\_minseplength}), or too close to borders (threshold \texttt{edge\_minborderdist}) are discarded.

\subsubsection{Band-based Separator Detection}
\label{sec:band-based-algorithm}

\begin{figure}
\includegraphics[width=\textwidth]{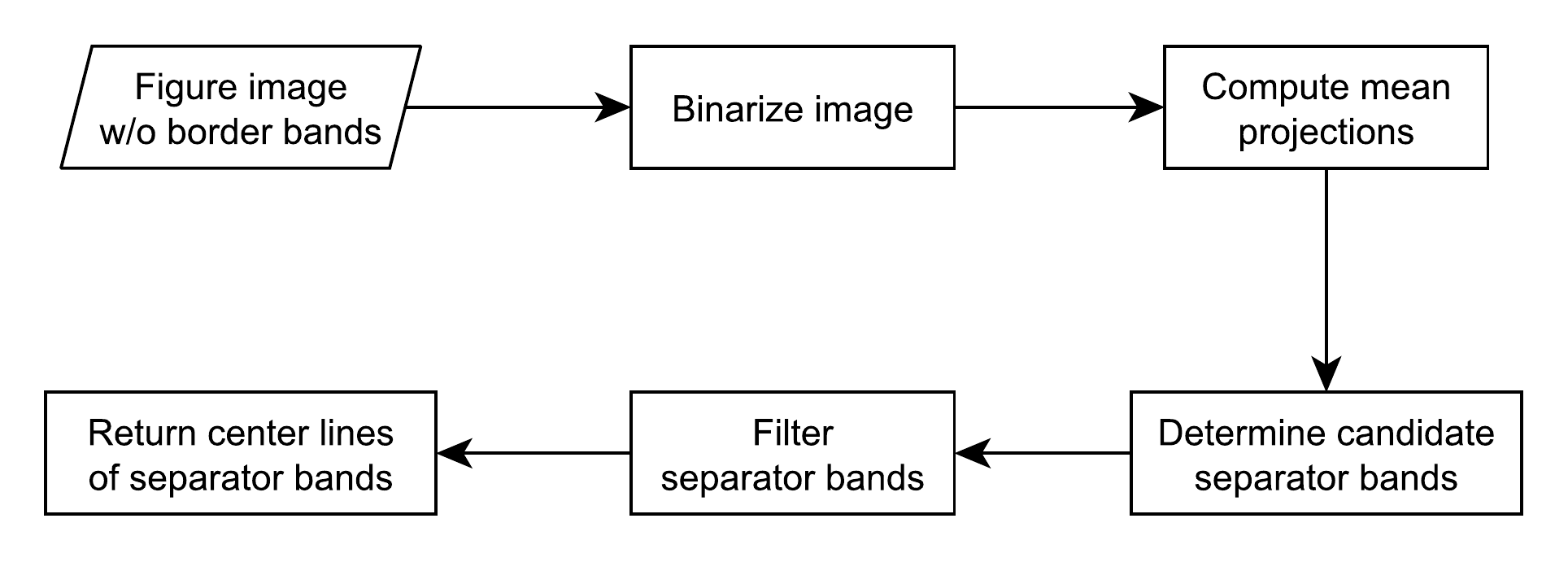}
\caption{Band-based separator line detection.}
\label{fig:band-based-algorithm}
\end{figure}

The band-based separator detection algorithm aims at locating homogeneous rectangular areas covering the full width or height of the image, which we call \emph{separator bands}. Since this algorithm is intended primarily for gray-scale illustration images with light background, we assume that separator bands are white or light gray. The algorithm consists of four steps illustrated in Fig.~\ref{fig:band-based-algorithm}: (1) image binarization, (2) computation of mean projections, (3) identification and (4) filtering of candidate separator bands.

Initially, we binarize the image using the mean intensity value as a threshold. We then compute mean projections along direction~$\theta$ (vertical or horizontal), that is, the mean value of each line of pixels in this direction. A resulting mean value will be 1 (white) if and only if the corresponding line contains only white pixels. Candidate separator bands are then determined by identifying maximal runs of ones in the vector of mean values that respect a minimal width threshold (\texttt{band\_minsepwidth}). They are subsequently filtered using a regularity criterion similar to Hough peak selection (see Section~\ref{sec:edge-based-algorithm}), this time using distance variance threshold \texttt{band\_maxdistvar}. Finally, selected bands that are close to the image border (threshold \texttt{band\_minborderdist}) are discarded, and the center lines of remaining bands are returned as separator lines.

\subsection{Chained Classification and Separation}
\label{sec:cfc-cfs-chain}

\begin{figure}
\centering
\includegraphics[width=\textwidth]{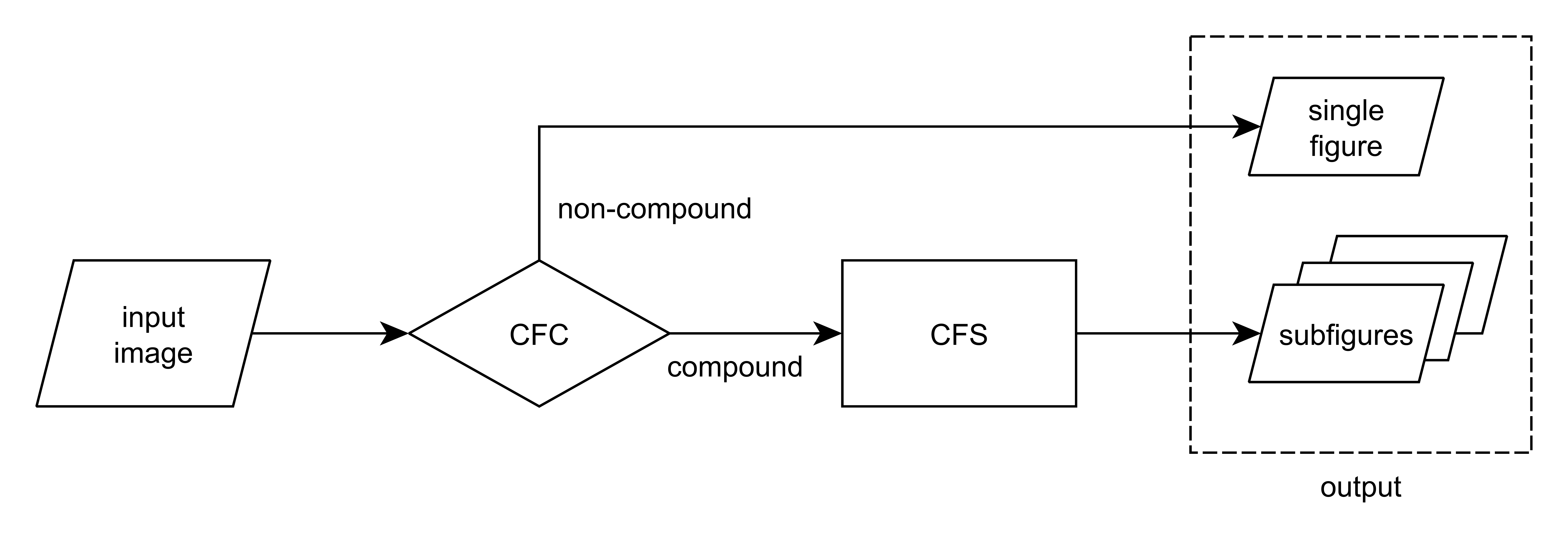}
\caption{Process chain consisting of compound figure classifier (CFC) and compound figure separation (CFS).}
\label{fig:cfc-cfs-chain}
\end{figure}

Processing compound figures in a collection of scientific articles is expected to happen in a two-stage process as illustrated in Fig.~\ref{fig:cfc-cfs-chain}: (1) all article images are classified as \emph{compound} or \emph{non-compound} by applying a compound figure classifier (CFC); (2) the predicted \emph{compound} images are then processed by a compound figure separation (CFS) algorithm to obtain subfigures. The resulting set of subfigures and predicted \emph{non-compound} figures can then be used for further application-specific processing (e.g. content-based indexing for retrieval). We are therefore interested in evaluating and improving the effectiveness of the \emph{CFC-CFS process chain}, i.e. the quality of obtained subfigures and non-compound figures with respect to a gold standard and evaluation procedure (see Section~\ref{sec:evaluation-methods}).

Our proposed method for evaluating the effectiveness of the CFC-CFS chain will be described in Section~\ref{sec:evaluation-methods}. A guiding principle for improving the CFC-CFS chain is derived from consideration of the loss of effectiveness caused by different types of CFC errors: \emph{false negatives} (compound figures classified as non-compound) may result in a larger loss than the same number of \emph{false positives} (non-compound figures classified as compound), because false negatives are not processed by CFS and hence all contribute to the loss of effectiveness. On the other hand, there is a chance that false positives are not divided into subfigures by CFS (because it does not detect separation lines), and such instances of false positives will not degrade effectiveness of the CFC-CFS chain. Effectiveness can therefore be optimized on a validation set by biasing CFC decisions towards the \emph{compound} class. However, this is easy to achieve only for CFC algorithms that deliver predicted class probabilities, like logistic regression, but not for SVM.

The different importance of misclassifications of a binary classifier depending on true classes can be expressed by a $2\times 2$ misclassification loss matrix (Eq.~\eqref{eq:loss-matrix}) \cite{Bishop2006ch1}. Rows correspond to true classes and columns to predicted classes, where in the case of CFC the first row or column is assigned to class \emph{non-compound} ($C_0$) and the second row or column to class \emph{compound} ($C_1$). The entries of loss matrix (Eq.~\eqref{eq:loss-matrix}) denote the fact that misclassification of true compound figures incurs a loss that is by a factor of $\alpha$ larger than that of misclassification of true non-compound figures (if $\alpha > 1$). If the classifier is able to predict class probabilities $p(C_k|x)$ for a given image~$x$, the decision of the classifier can be optimized with respect to expected misclassification loss $E_k(x)$ (Eq.~\eqref{eq:expected-loss}): image $x$ is assigned to class $C_k$ that minimizes $E_k(x)$ ($k=0$ or $k=1$). For the special form of loss matrix given in (Eq.~\eqref{eq:loss-matrix}), this criterion reduces to a simple threshold on class probability $p(C_1|x)$: image $x$ is assigned to class $C_1$ if and only if Eq.~\eqref{eq:decision-criterion} holds. The parameter $\alpha$ can be selected by optimizing effectiveness of the CFC-CFS process chain on a validation set.

\begin{eqnarray}
	L &=& \begin{pmatrix} 0 & 1\\ \alpha & 0\end{pmatrix} \label{eq:loss-matrix}\\
	E_k(x) &=& \sum_{i} L_{ik}\, p(C_i|x) \label{eq:expected-loss} \\
	p(C_1|x) &\geq& \frac{1}{1+\alpha} \label{eq:decision-criterion}
\end{eqnarray}

\section{Experiments and Results}
\label{sec:experiments}

We evaluate our approach on separate datasets for CFC, CFS, and the CFC-CFS process chain, which are described in Section~\ref{sec:datasets}. As there is no agreement on a standard evaluation protocol for CFS in the research community yet, we use two different evaluation procedures,  described in Section~\ref{sec:evaluation-methods}. Additionally, we propose to slightly extend existing CFS evaluation protocols in order to apply them to CFC-CFS chains. Evaluation results for CFC, CFS, and CFC-CFS chain are presented and discussed in Sections~\ref{sec:cfc-experiments}, \ref{sec:cfs-experiments}, and \ref{sec:cfc-cfs-experiments}, respectively.

\subsection{Datasets}
\label{sec:datasets}

We used several datasets to train and evaluate the different components of our approach in our experiments. All of them were derived from the dataset of about 75,000 biomedical articles used for ImageCLEF medical tasks since 2012 \cite{Kalpathy-Cramer2015}. Those articles were retrieved from PubMed Central\footnote{\url{http://www.ncbi.nlm.nih.gov/pmc/}} by selecting open access journals that allow for free redistribution of data. The articles of the ImageCLEF dataset contain about 300,000 images of unconstrained modalities (biomedical images, diagrams, charts, photographs, etc.) and subfigure structure (\emph{compound} and \emph{non-compound} images).

A subset of about 21,000 images used for the ImageCLEF 2015 medical tasks \cite{Herrera2015} formed the basis for most datasets used in our experiments, namely all datasets labeled \emph{ImageCLEF} in Table~\ref{tab:datasets}.
The CFC training dataset provided by ImageCLEF task organizers contained some erroneous samples (23 images had contradicting annotations), which have been removed from the training set. Table~\ref{tab:datasets} refers to the cleaned CFC training set only. The CFC dataset consists of 59\% compound images (CO), both in training and test subsets, providing reasonable conditions for training and evaluating a binary classifier. A similar split of classes is present in the modality classification (MC) datasets, which are used to train and evaluate the binary classifier for illustrations (ILL) (see Section~\ref{sec:illustration-classifier}).

\begin{table}
\caption{Datasets used in our experiments. CFC~= compound figure classification, CFS~= compound figure separation, MC~= modality classification; CO~= compound, ILL~= illustration.}
\label{tab:datasets}
\footnotesize
\begin{tabularx}{\textwidth}{Xrlrl}
\hline
\multirow{2}{*}{Dataset}             & \multicolumn{2}{c}{Training}   & \multicolumn{2}{c}{Test} \\
\cline{2-5}
                                     & Images & Annotations           & Images & Annotations \\
\hline
ImageCLEF CFC                        & 10387  & 6121 CO (59\%)  & 10434  & 6144 CO (59\%) \\
ImageCLEF CFS                        & 3403   & 14531 subfigures      & 3381   & 12789 subfigures \\
NLM CFS                              &        &                       & 380    & 1656 subfigures \\
ImageCLEF MC first                   & 1071   & 607 ILL (57\%) & 497  & 261 ILL (53\%) \\
ImageCLEF MC majority                & 895    & 514 ILL (57\%) & 428  & 243 ILL (57\%) \\
ImageCLEF MC unanimous               & 867    & 508 ILL (59\%) & 398  & 226 ILL (57\%) \\
ImageCLEF MC greedy                  & 1071   & 712 ILL (66\%) & 497  & 325 ILL (65\%) \\
CFC-CFS                              & 6806   & 17934 subfigures      & 6752   & 16154 subfigures \\
\hline
\end{tabularx}
\normalsize
\end{table}

The MC datasets were derived from the dataset of the ImageCLEF 2015 multi-label image classification task~\cite{Herrera2015}. The images are provided with one or more labels of 29~classes (organized in a class hierarchy), which have been mapped to two meta classes: the \emph{illustration} meta class comprises all ``general biomedical illustration" classes except for chromatography images, screenshots, and non-clinical photos. These classes and all classes of diagnostic images have been assigned to the \emph{non-illustration} meta class. About 36\% of the images in the training set are labeled with multiple classes, corresponding to compound images. Training and evaluation of the illustration classifier (Section~\ref{sec:illustration-classifier}) requires mapping the set of labels of a given image to a single meta class. We implemented four mapping strategies that first assign each image label to the \emph{illustration} or \emph{non-illustration} meta class, and then operate differently on the list $L$ of meta labels associated with a given image: (1) the \emph{first} strategy simply assigns the first meta label of $L$ to the image; (2) the \emph{majority} strategy selects the meta label occurring most often in $L$, dropping the image from the dataset if both meta labels occur equally often; (3) the \emph{unanimous} strategy only assigns a meta label to the image if all meta labels in $L$ are equal, otherwise the image is dropped from the dataset; and (4) the \emph{greedy} strategy maps an image to the \emph{illustration} label if $L$ contains at least one such meta label, otherwise the image is assigned the \emph{non-illustration} label. Note that \emph{majority} and \emph{unanimous} strategies discarded up to 20\% of images in the original dataset. Whereas \emph{majority} and \emph{unanimous} mapping strategies are expected to improve classification accuracy, the \emph{greedy} strategy aims at increasing CFS effectiveness based on the assumption that a compound image containing an illustration subfigure is more likely to have separator bands than separator edges.

A research group at the U.S. National Library of Medicine (NLM) had created a dataset to evaluate their CFS approach (and related algorithms) \cite{Apostolova2013} well before the first CFS task at ImageCLEF was issued in 2013. This dataset contains 400 images and 1764 ground-truth subfigures and hence is substantially smaller than the ImageCLEF CFS test dataset. Moreover, it shares 20~images with the training set and  27~images with the test set of the ImageCLEF CFS dataset. The reason for the non-empty intersection of these datasets is that the NLM dataset was sampled from a set of 231,000 article images used at ImageCLEF 2011, which was extended later to the ImageCLEF dataset mentioned at the beginning of this section. Since we used the ImageCLEF CFS training set for parameter optimization, we removed the 20~images in the intersection from the NLM dataset for our experiments. The resulting reduced dataset is listed in Table~\ref{tab:datasets} as \emph{NLM CFS } dataset.

For evaluation of the CFC-CFS process chain, we extended the ImageCLEF CFS test dataset (3381 images) with the same number of non-compound images sampled at random from the ImageCLEF CFC test dataset. After removing five images that occurred in both portions of this dataset\footnote{Ideally, the intersection should be empty, because the CFS dataset should contain only compound images. However, manual inspection of images in the intersection revealed that both CFS and CFC datasets contain errors and that the distinction between compound and non-compound images is not always clear.}, a test dataset with 6752 images was obtained. In a similar manner, a validation set of 6806 images was constructed from ImageCLEF CFS and CFC training datasets (appearing as ``training set'' in the last line of Table~\ref{tab:datasets}). Non-compound images of the CFC-CFS dataset were annotated with a single subfigure covering the entire image, as explained in Section~\ref{sec:chain-evaluation}.

\subsection{Evaluation Methods}
\label{sec:evaluation-methods}

While evaluation of classification algorithms is a well-studied problem \cite{Caruana2006,Hastie2009ch7,Kou2012,Mitchell1997ch5,Smith-Miles2009}, evaluation of compound figure separation has been addressed by two different ad-hoc procedures only \cite{Apostolova2013,Herrera2013}. Both evaluation procedures first determine which detected subfigures of a given compound image are correct (\emph{true positive}) with respect to ground-truth subfigures, and then compute an evaluation measure from the number of true positive subfigures over the dataset. However, the way by which true positive subfigures are determined, and which evaluation measures are calculated, differs between the two proposed evaluation procedures, which are described in detail in the following section. In Section~\ref{sec:chain-evaluation} we propose to also apply CFS evaluation methods to measure the effectiveness of the CFC-CFS process chain.

\subsubsection{CFS Evaluation}
\label{sec:cfs-evaluation}

To describe the evaluation protocols in detail, we introduce the following notation. Without loss of generality, we assume that a subfigure is represented by rectangular area $R$ (bounding box) within an image, and denote its area size (number of contained pixels) by $|R|$. For a given compound figure, let $\{G_i\,|\,i\in I\}$ be the set of ground-truth subfigures, and $\{F_j\,|\,j\in J\}$ the set of subfigures detected by the CFS algorithm that should be evaluated. Note that the overlap area $G_i\cap F_j$ between subfigures is again a rectangle (or empty). The two evaluation protocols employ different definitions of the \emph{overlap ratio} between $G_i$ and $F_j$, given in Equations \eqref{eq:overlap-g} and \eqref{eq:overlap-f}. $\rho_{ij}^G$ is the overlap ratio with respect to ground-truth subfigure $G_i$, $\rho_{ij}^F$ calculates the ratio with respect to detected subfigure $F_j$.

\begin{eqnarray}
\rho_{ij}^G &=& \frac{|G_i\cap F_j|}{|G_i|} \label{eq:overlap-g}\\
\rho_{ij}^F &=& \frac{|G_i\cap F_j|}{|F_j|} \label{eq:overlap-f}
\end{eqnarray}

The evaluation procedure used for ImageCLEF CFS tasks \cite{Herrera2013} iterates over ground-truth subfigures $G_i$ and, for a given $G_i$, looks for a detected subfigure $F_j$ with maximal overlap $\rho_{ij}^F$. $F_j$ is associated with $G_i$ if $\rho_{ij}^F > 2/3$ and if $F_j$ has not already been associated with a different ground-truth subfigure. The result is a set of one-to-one associations between ground-truth subfigures and detected subfigures, which are regarded as true positives. Note that although the set of associations may depend on the order of iterations over $G_i$, the number $C$ of these associations does not. Accuracy can therefore be defined per compound figure as $C/\max(N_G,N_D)$, where $N_G$ and $N_D$ are the numbers of ground-truth and detected subfigures, respectively. Accuracy on the test set is the average of accuracy values computed for each compound figure.

The authors of the NLM CFS dataset \cite{Apostolova2013} (see Section \ref{sec:datasets}) used a different criterion to determine true positive subfigures. A detected subfigure $F_j$ is considered true positive if and only if there is a ground-truth subfigure $G_i$ with $\rho_{ij}^G > 0.75$ and $\rho_{kj}^G < 0.05$ for all other ground-truth subfigures $G_k$. That is, subfigure $F_j$ has a notable overlap with one ground-truth subfigure only. Given the total number $N$ of ground-truth subfigures in the dataset, the total number $D$ of detected subfigures, and the number $T$ of detected true positive subfigures, the usual definitions for classifier evaluation measures can be applied to obtain precision $P$, recall $R$, and $F_1$ measure, see Eq. \eqref{eq:precision-apostolova}. Note that accuracy is not well-defined in this setting, because the number of negative results (not detected arbitrary bounding boxes) is theoretically unlimited.

\begin{equation}\label{eq:precision-apostolova}
P = \frac{T}{D}\enspace,\quad R = \frac{T}{N}\enspace,\quad F_1 = \frac{2*P*R}{P+R}\enspace .
\end{equation}

\begin{figure}
\begin{tabular}{ll}
(a)\hspace*{8mm} & \includegraphics[valign=c, width=0.7\textwidth]{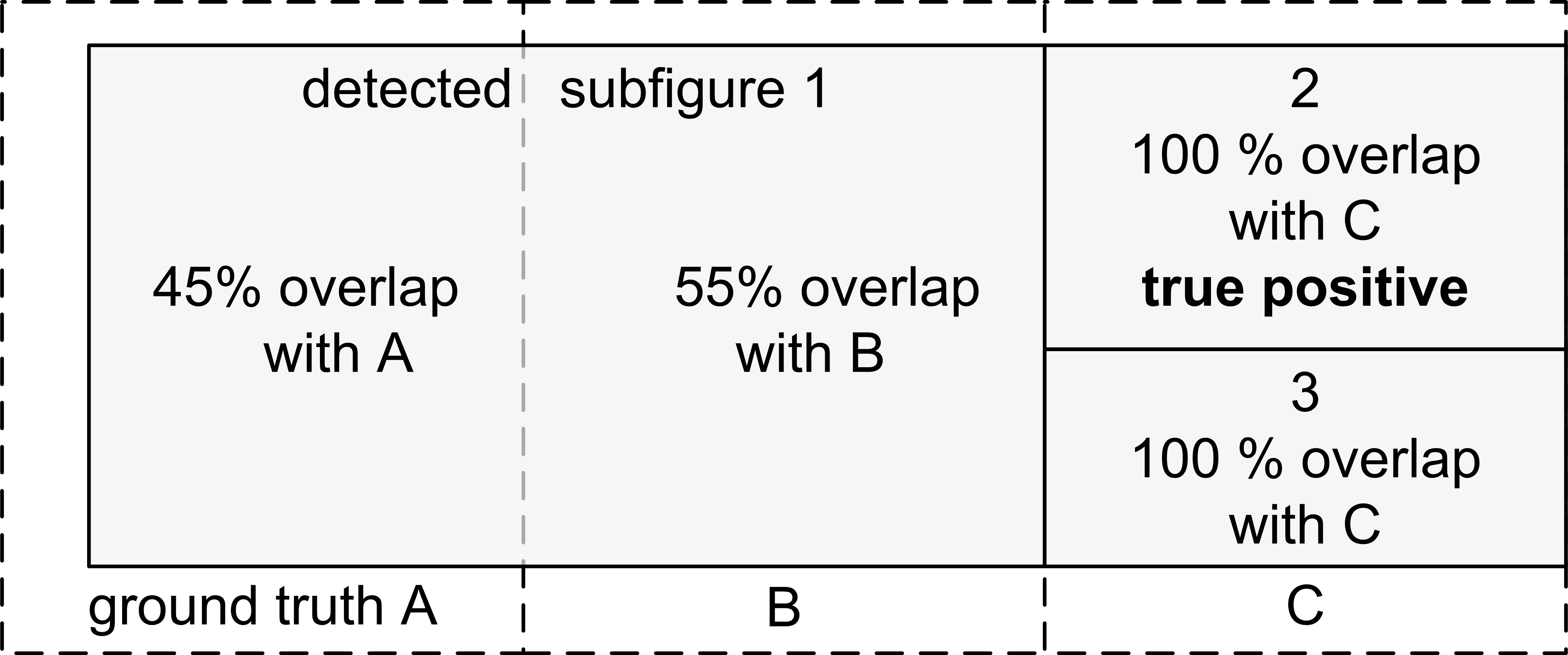}\\
& \\
(b) & \includegraphics[valign=c, width=0.7\textwidth]{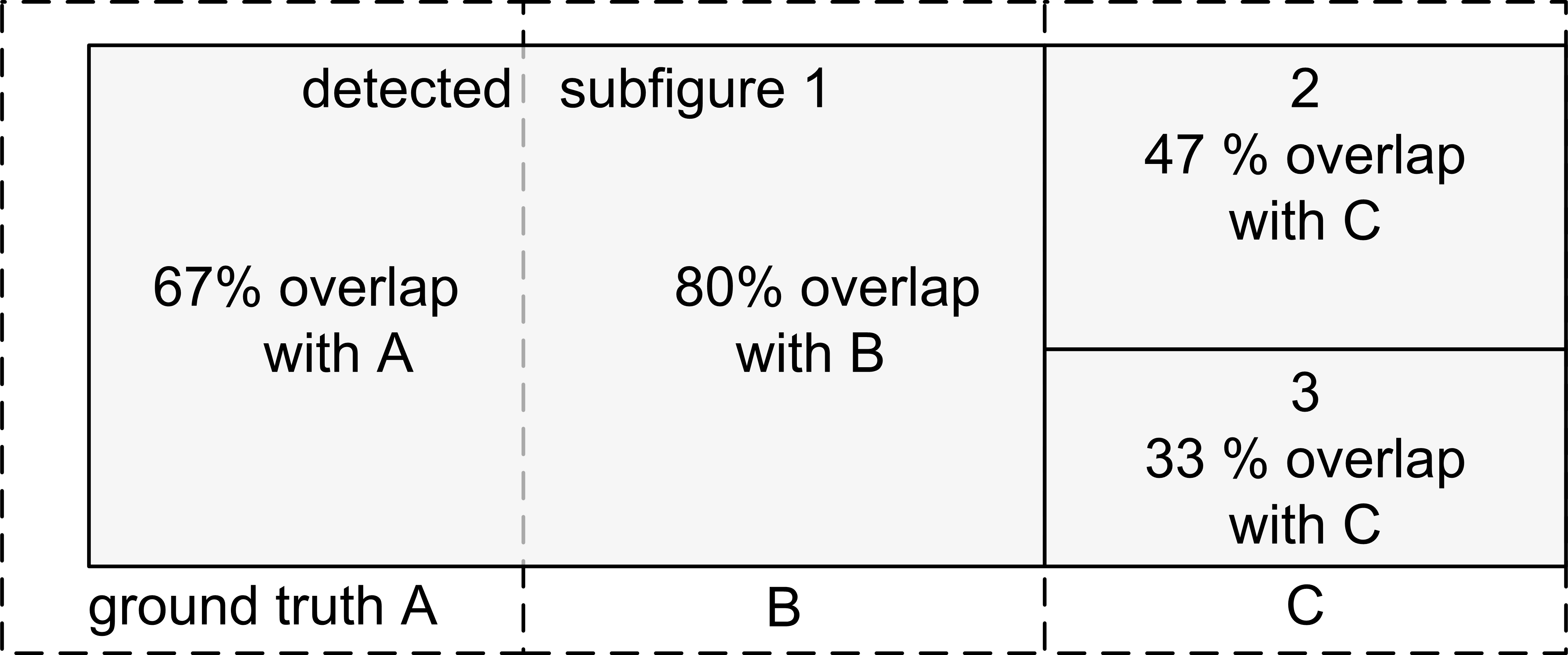}\\
\end{tabular}
\caption{Determination of true positive detected subfigures by (a) ImageCLEF and (b) NLM CFS evaluation procedures.}
\label{fig:cfs-evaluation}
\end{figure}

Figure \ref{fig:cfs-evaluation} illustrates two different ways of determining \emph{true positive} detected subfigures for an example compound figure, which consists of three ground-truth subfigures: A, B, and C. We assume that a hypothetical CFS algorithm, given this compound figure as input, produced three subfigures -- indicated as subfigures 1, 2, and 3 -- at its output. In Figure \ref{fig:cfs-evaluation}, the resulting detected subfigures appear on the foreground, partially overlapping the three ground-truth subfigures in the background. The ImageCLEF and NLM evaluation protocols for this case will result in two different assessments, as follows:

\begin{itemize}
\item
Figure \ref{fig:cfs-evaluation} (a): The ImageCLEF evaluation procedure considers only one of subfigures~2 or~3 as true positive, depending on which of them gets associated first with~C. Note that both overlap ratios $\rho_{C2}^F$ and $\rho_{C3}^F$ are 100\%. Subfigure~1, however, is regarded as false positive, because its overlap ratio, according to definition \eqref{eq:overlap-f}, with any ground-truth subfigure does not exceed 2/3. The resulting accuracy is therefore 1/3, since only one of the three detected subfigures qualifies as true positive.

\item
Figure \ref{fig:cfs-evaluation} (b): The NLM evaluation procedure, on the other hand, determines that all detected subfigures should be considered false positives, because for subfigures~2 and~3 the overlap ratio, according to definition \eqref{eq:overlap-g}, with any ground-truth subfigure is too small (i.e., less than 75\%), and subfigure~1 overlaps with two ground-truth subfigures (A and B) by at least 5\%.
\end{itemize}

\subsubsection{CFC-CFS Chain Evaluation}
\label{sec:chain-evaluation}

We propose to apply the CFS evaluation methods described in the previous section to the output of the CFC-CFS process chain (Section~\ref{sec:cfc-cfs-chain}). Because CFS test datasets contain only compound figures, but the dataset for CFC-CFS chain evaluation also includes non-compound figures (Section~\ref{sec:datasets}), we need to extend CFS evaluation procedures by a convention to represent non-compound figures. We adopt the obvious solution to consider non-compound figures as ``compound figures with a single subfigure'' and represent each of them by a bounding box covering the entire image. This extension needs to be implemented in three different places of the evaluation procedure: (1) for ground-truth annotation, (2) for images classified as \emph{non-compound} by CFC, and (3) for images classified as \emph{compound} that are not divided into subfigures by CFS (because it does not detect proper separator lines).

Unmodified CFS evaluation algorithms can then be applied to the output of the CFC-CFS chain. Note that the ImageCLEF evaluation algorithm will assign 100\% accuracy for true non-compound images only if there is exactly one ``detected'' subfigure in the CFC-CFS output, no matter what the bounding boxes are. Similarly, the NLM evaluation algorithm will find at most one true positive subfigure in a true non-compound image, but in this case the area of the ``detected'' bounding box is relevant (it must cover at least 75\% of the entire image).

\subsection{Compound Figure Classifier}
\label{sec:cfc-experiments}

We used the ImageCLEF CFC dataset (Section~\ref{sec:datasets}) to train and evaluate the various combinations of feature sets and classifier algorithms described in Section~\ref{sec:compound-fig-classifier}. More specifically, we trained all three classifiers on 40 feature sets created by instantiating the 10 feature sets listed in Table~\ref{tab:cfc-features-sets} for four values of $k$ (4, 8, 16, and 32). The quantization parameters were kept constant as $p=5$, $q=8$, and $h=3$, as these values gave good classification performance in preliminary experiments. To enable a fair comparison with SVM, the logistic regression classifier used a probability threshold of 0.5, corresponding to a symmetric misclassification loss matrix (Eq. \eqref{eq:loss-matrix}) with $\alpha=1$.


Results of CFC experiments are presented in Table~\ref{tab:cfc-results}. From the 120 combinations of classifier algorithm, feature sets, and number $k$ of spatial bins that were tested in experiments, we report only the best three and the worst results -- separated by a dashed line in Table~\ref{tab:cfc-results} --  for each classifier algorithm with respect to accuracy.

\begin{table}
\caption{Evaluation results of compound figure classifier on ImageCLEF CFC test set. From the 120 tested combinations of classifier algorithm, feature set, and number $k$ of spatial bins, only the best three and the worst result for each classifier algorithm are reported. LogReg = logistic regression, SVM = support vector machine.}
\label{tab:cfc-results}
\centering
\begin{tabular}{llrrrr}
\hline
\textbf{Classifier} & \textbf{Feature Set} & $k$ & \textbf{Accuracy\%} & \textbf{FP\%} & \textbf{FN\%}\\
\hline
LogReg & 134 & 16 & 76.9 & 16.9 & 6.2 \\
LogReg & 434 & 8  & 76.6 & 18.2 & 5.2 \\
LogReg & 434 & 16 & 76.6 & 17.7 & 5.7 \\
\hdashline
LogReg & 011 & 4  & 61.3 & 8.5 & 30.2 \\
\hline
linear SVM & 134 & 16 & 76.9 & 14.6 & 8.6 \\
linear SVM & 434 & 8  & 76.8 & 16.6 & 6.7 \\
linear SVM & 434 & 16 & 76.5 & 15.9 & 7.6 \\
\hdashline
linear SVM & 222 & 4  & 63.9 & 25.9 & 10.2 \\
\hline
kernel SVM & 034 & 4 & 75.5 & 20.4 & 4.1 \\
kernel SVM & 444 & 4 & 75.3 & 20.8 & 3.9 \\
kernel SVM & 434 & 4 & 74.2 & 23.0 & 2.9 \\
\hdashline
kernel SVM & 666 & 32 & 59.0 & 41.0 & 0.0 \\
\hline
\end{tabular}
\end{table}

Results indicate that feature set 434 achieves good classification performance for all three tested classifier algorithms with a rather low dimensionality of 96 (see Table~\ref{tab:cfc-features-sets}). Feature set 134 (with 224 dimensions) with $k=16$ spatial bins showed the  same accuracy (76.9\%) for both linear classifiers, becoming the best overall performer in both cases. The surprisingly low classification performance of kernel SVM is probably due to underfitting caused by default SVM hyperparameters; both box constraint $C$ and standard deviation $\sigma$ of the radial basis function (RBF) kernel were kept at the default value~1.


Remarkably, the false positive rate of all well-performing classifiers in Table~\ref{tab:cfc-results} is systematically higher than the false negative rate. This can be explained by two possible causes: first, the training set is slightly imbalanced (59\% compound images), which may cause the classifier to decide in favor of the \emph{compound} class in uncertain cases; second, the feature sets used for CFC produce a denser spatial distribution of non-compound images in the feature space than for compound ones, reinforcing the imbalanced training effect. In fact, the CFC features described in Section~\ref{sec:compound-fig-classifier} have been designed to capture the existence of separators between subfigures. If such separators do not exist, feature values may exhibit a low variance across different images.

Compared to the best CFC run using visual-only features submitted to ImageCLEF 2015 by Wang et al. \cite{Wang2015}, which achieved 82.8\% accuracy on the same dataset, our results are inferior by a margin of about 6\%. However, as the approach of Wang et al. essentially employs a CFS algorithm (connected component analysis and band separator detection), we suppose that our CFC method has significant advantages with respect to efficiency for online classification. Extraction of the 111 feature set, which is the most complex of our proposed feature sets, took 81~milliseconds per image on average (excluding reading the image file from disk) using a MATLAB implementation on an Intel E8400 CPU operated at 3~GHz. This execution time corresponds to a processing rate of 12.3~images per second.

\subsection{Compound Figure Separation}
\label{sec:cfs-experiments}


Our CFS approach is evaluated mainly on the ImageCLEF CFS dataset (Section~\ref{sec:datasets}) using the ImageCLEF evaluation procedure (Section~\ref{sec:evaluation-methods}). The internal parameters of our CFS algorithm, including implementation options of the illustration classifier, are optimized on the training portion of the dataset as described in Section~\ref{sec:param-optimization}, prior to evaluating CFS performance on the test dataset. To analyze the effectiveness of the illustration classifier for CFS, we also report results for different classifier implementation options obtained by keeping these options constant during parameter optimization.

Moreover, we consider a variant of the proposed CFS algorithm in which the illustration classifier has been replaced by a binary random decision unit, which predicts that a given input image is an \emph{illustration} with probability~$p$. For $p=0$, the CFS algorithm will always use edge-based separator detection, and for $p=1$ band-based separator detection will be applied to every input image. The rationale for choosing $p$ as the actual \emph{illustration} decision rate of the classifier on the test dataset is to allow a fair comparison between the ``random decision'' variant and the proposed CFS algorithm, which should allow us to quantify the utility of the illustration classifier in our CFS approach.

\begin{figure}
\centering
\includegraphics[width=\textwidth]{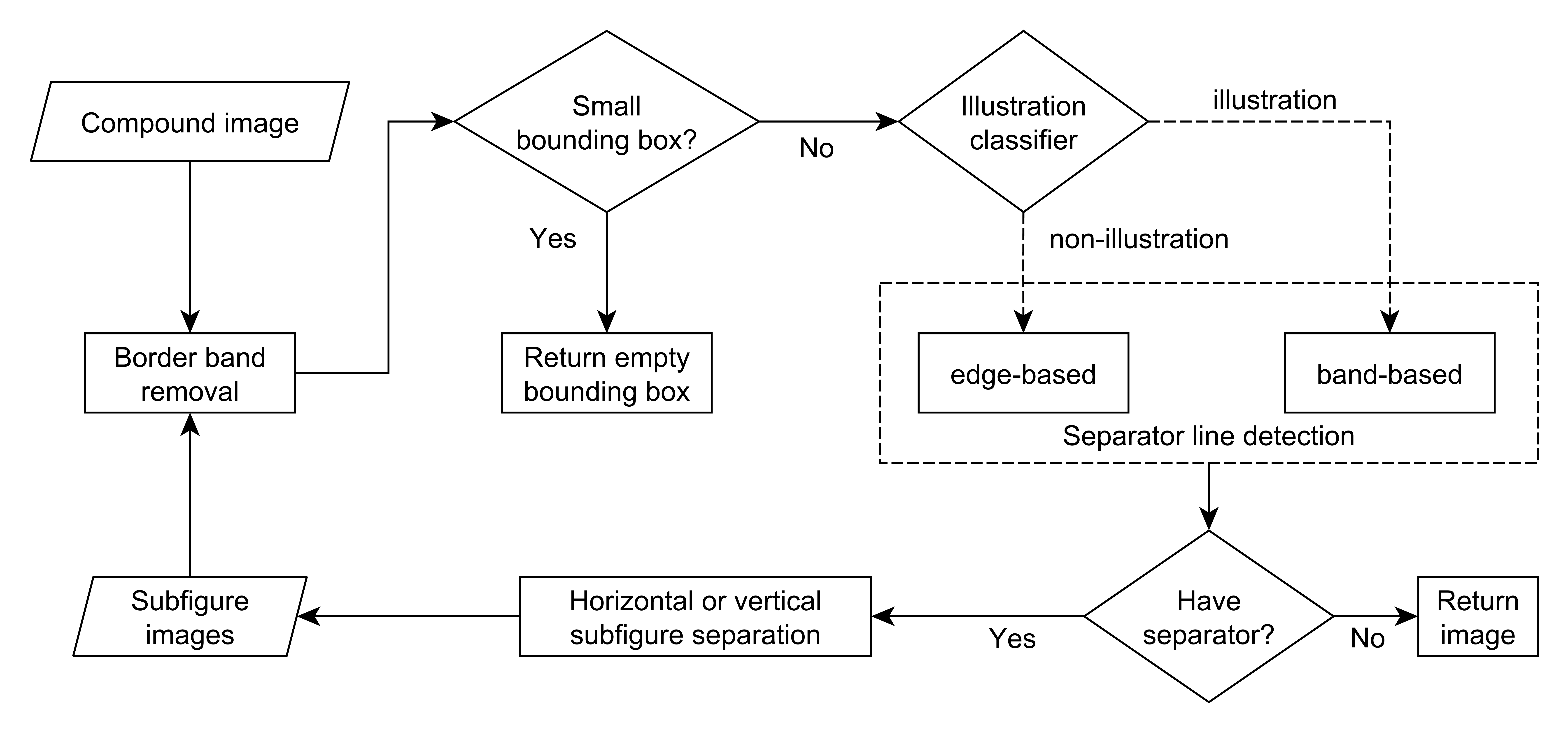}
\caption{Variant of proposed CFS algorithm that applies the illustration classifier to every detected subfigure prior to splitting it further.}
\label{fig:main-algorithm-variant}
\end{figure}

The proposed CFS algorithm applies the illustration classifier once to each input image and reuses the classifier's decision in all recursive invocations of the separator detection module (see Fig.~\ref{fig:recursive-algorithm}). To answer the question whether applying the classifier anew for each recursive invocation improves CFS performance, we also consider this algorithmic variant in our experiments, depicted in Fig.~\ref{fig:main-algorithm-variant}.

To enable comparison with other CFS approaches in the literature, we further evaluate our approach on the NLM dataset using the evaluation procedure proposed by its authors (see Section~\ref{sec:evaluation-methods}). By using the same parameter values obtained by optimization on the ImageCLEF training set, CFS results on the NLM dataset provide additional information about the generalization ability of our CFS algorithm.

\subsubsection{Parameter Optimization}
\label{sec:param-optimization}

The proposed CFS algorithm takes 17 internal parameters listed in Table~\ref{tab:parameters}.
Initial parameter values were chosen manually by looking at the results produced for a few training images. They were used during participation in ImageCLEF 2015 \cite{Taschwer2015}. For parameter optimization, the CFS algorithm was evaluated for various parameter combinations on the ImageCLEF 2015 CFS training dataset (3,403 compound images, 14,531 ground-truth subfigures) using the evaluation tool provided by ImageCLEF organizers. Due to the number of parameters and the run time of a single evaluation run (about 17~minutes), a grid-like optimization evaluating all possible parameter combinations in a certain range was not feasible. Instead, we applied a hill-climbing optimization strategy to locate the region of a local maximum and then used grid optimization in the neighborhood of this maximum.

\begin{table}
\caption{Internal parameters of proposed CFS algorithm. Initial parameter values were used during ImageCLEF 2015 participation \cite{Taschwer2015}, optimal values were obtained by parameter optimization on the ImageCLEF 2015 CFS training dataset. Parameters marked by * use units of image width, height, or area, depending on the parameter and processing direction (horizontal or vertical).}
\label{tab:parameters}
\begin{center}
\begin{tabular}{lrr>{\raggedright\arraybackslash}p{50mm}}
\hline
\textbf{Parameter} & \textbf{Initial} & \textbf{Optimal} & \textbf{Meaning} \\
\hline
\multicolumn{4}{l}{\textbf{Main algorithm}}\\
\texttt{classifier\_model} & first & greedy 
	& first, majority, unanimous, or greedy (see Section~\ref{sec:illustration-classifier})	\\
\texttt{decision\_threshold} & 0.5 & 0.1 & minimal illustration class probability to decide in favor of band-based separator detection\\
\texttt{mindim} & 50 & 200 
	& minimal image dimension (pixels) to apply separator detection to\\
\texttt{elim\_area} & 0 & 0.03
	& area threshold to eliminate small bounding boxes*\\
\hline
\multicolumn{4}{l}{\textbf{Edge-based separator detection}}\\
\texttt{edge\_maxdepth} & 10 & 10 & maximal recursion depth\\
\texttt{edge\_sobelthresh} & 0.05 & 0.02
	& threshold for Sobel edge detector\\
\texttt{edge\_houghratio\_min} & 0.25 & 0.2
	& minimal ratio of Hough values for peak selection\\
\texttt{edge\_houghratio\_base} & 1.2 & 1.5
	& base of recursion depth dependency for Hough peak selection\\
\texttt{edge\_maxdistvar} & 0.0001 & 0.1
	& maximal variance of separator distances for regularity criterion*\\
\texttt{edge\_gapratio} & 0.2 & 0.3 
	& gap threshold for edge filling*\\
\texttt{edge\_lenratio} & 0.05 & 0.03
	& length threshold for edge filling*\\
\texttt{edge\_minseplength} & 0.7 & 0.5
	& minimal separator length*\\
\texttt{edge\_minborderdist} & 0.1 & 0.05
	& minimal distance of separators from border*\\
\hline
\multicolumn{4}{l}{\textbf{Band-based separator detection}}\\
\texttt{band\_maxdepth} & 2 & 4
	& maximal recursion depth\\
\texttt{band\_minsepwidth} & 0.03 & 0.0001
	& minimal width of separator bands*\\
\texttt{band\_maxdistvar} & 0.0003 & 0.2
	& maximal variance of separator distances for regularity criterion*\\
\texttt{band\_minborderdist} & 0.1 & 0.01
	& minimal distance of separators from border*\\
\hline
\end{tabular}
\end{center}
\end{table}

More precisely, we defined up to five different values per parameter, including the initial values, on a linear or logarithmic scale, depending on the parameter. Then a set of parameter combinations was generated where only one parameter was varied at a time and all other parameters were kept at their initial values, resulting in a feasible number of parameter combinations to evaluate (linear in the number of parameters). After measuring accuracy on the training set, the most effective value of each parameter was chosen as its new \emph{optimal} value. For parameters whose optimal values differed from the initial ones, the range was centered around the optimal value. Other parameters were fixed at their latest value. The procedure was repeated until accuracy improved by no more than~5\%, which happened after three iterations. Finally, after sorting parameter combinations by achieved accuracy, the five most effective parameters were chosen for grid optimization, where only two ``nearly optimal" values (including the latest optimal value) per parameter were selected.

The effect of parameter optimization was surprisingly strong: whereas the initial parameter configuration achieved an accuracy of 43.5\% on the training set, performance increased to 84.5\% after hill-climbing optimization, and finished at 85.5\% after grid optimization.

\subsubsection{Evaluation on ImageCLEF Dataset}
\label{sec:evaluation-imageclef}

Experimental results are shown in Table~\ref{tab:evaluation-results}. For comparison, we also included a previous version of our approach \cite{Taschwer2015} that did not use optimized parameters, and the best approach submitted to ImageCLEF 2015 (by NLM). We evaluated the proposed algorithm with optimized parameters (see Section~\ref{sec:param-optimization}) and with different implementations and feature sets for the illustration classifier, as described in Section~\ref{sec:illustration-classifier}. Because logistic regression using \emph{simple2} features was found to be most effective by parameter optimization when trained on the \emph{greedy} set, we focused on this training set when evaluating other classifier implementations. Internal SVM parameters were optimized on the entire ImageCLEF 2015 multi-label classification test dataset (see Section~\ref{sec:evaluation-classifier}) to maximize classification accuracy. The optimized \texttt{decision\_threshold} parameter for deciding between edge-based and band-based separator detection is effective only for logistic regression classifiers, because SVM predictions do not provide class probabilities. To confirm the effectiveness of the illustration classifier, we also included results for algorithm variants where the classifier has been replaced by a random decision selecting band-based separator detection with probability~$p$. The value $p=0.741$ corresponds to the decision rate of the most effective classifier (LogReg,simple11,greedy). $p=0$ and $p=1$ represent algorithms that always use edge-based or band-based separator detection, respectively. Finally, the algorithm variant \emph{SubfigureClassifier} applies the illustration classifier not only once per compound image, but also to each subimage during recursive figure separation (see Fig.~\ref{fig:main-algorithm-variant}).

\begin{table}
\caption{Experimental results on the ImageCLEF 2015 CFS test set. Illustration classifiers are described in Section~\ref{sec:illustration-classifier} (LogReg = logistic regression). BB denotes the percentage of images (or decisions*) where band-based separator detection was applied.}
\label{tab:evaluation-results} 
\begin{center}
\begin{tabular}{llcc}
\hline
\textbf{Algorithm} & \textbf{Classifier} & \textbf{BB \%} & \textbf{CFS Accuracy \%} \\
\hline
Previous \cite{Taschwer2015} & LogReg,simple2,first          &      & 49.4 \\
NLM \cite{Santosh2015}       & manual                        & 95.7 & 84.6 \\
Proposed                     & LogReg,simple2,first          & 61.6 & 84.2 \\
Proposed                     & LogReg,simple2,majority       & 61.1 & 84.1 \\
Proposed                     & LogReg,simple2,unanimous      & 61.8 & 84.2 \\
Proposed                     & LogReg,simple2,greedy         & 75.8 & 84.8 \\
Proposed                     & LogReg,simple11,greedy        & 74.1 & \textbf{84.9} \\
Proposed                     & SVM,simple2,greedy            & 58.6 & 83.5 \\
Proposed                     & SVM,simple11,greedy           & 60.3 & 83.5 \\
Proposed                     & SVM,CEDD,greedy               & 59.2 & 82.8 \\
Proposed                     & SVM,CEDD\_simple11,greedy     & 59.6 & 83.2 \\
Proposed                     & random,p=0.741                & 74.7 & 75.4 \\
Proposed                     & no classifier,p=0             &    0 & 58.0 \\
Proposed                     & no classifier,p=1             &  100 & 82.2 \\
SubfigureClassifier          & LogReg,simple11,greedy        & 60.1* & 84.0 \\
\hline
\end{tabular}
\end{center}
\end{table}

When comparing our results to NLM's approach, we note that the authors of \cite{Santosh2015} manually classified the test set into stitched (4.3\%) and non-stitched (95.7\%) images, whereas our approach uses automatic classification. Using band-based separator detection for all test images (no classifier, $p=1$) works surprisingly well (82.2\% accuracy), which can be explained by the low number of stitched compound images in the test set. On the other hand, using edge-based separator detection for all test images (no classifier, $p=0$) results in modest performance (58\% accuracy), which we attribute to a significant number of subfigures without rectangular borders (illustrations) in the test set. Selecting edge-based or band-based separator detection using the illustration classifier improved accuracy for all tested classifier implementations. In fact, it turned out to be effective to bias the illustration classifier towards band-based separator detection and apply edge-based separator detection only to high-confidence non-illustration images. This happened in two ways: by using the \emph{greedy} training set, and by optimizing the \texttt{decision\_threshold} parameter for the logistic regression classifier. This explains why best results were obtained by logistic regression classifiers trained on the \emph{greedy} training set.

To further analyze the effectiveness of separator detection selection, we partitioned the CFS test dataset into two classes according to decisions of the most effective CFS algorithm variant (LogReg,simple11,greedy) and evaluated detection results of this algorithm separately on the two partitions. Resulting accuracy values of 85.7\% on the edge-based partition and 84.6\% on the band-based partition show that the classifier was successful in jointly optimizing detection performance for both separator detection algorithms.

Our algorithm was implemented in MATLAB and executed on a PC with 8~GB RAM and an Intel E8400 CPU running at 3~GHz. The average total processing time per compound image was 0.3~seconds when an illustration classifier with \emph{simple} features was used, and 0.9~seconds when a classifier with CEDD features was applied. Note that the efficiency of other known approaches in the literature is either not documented \cite{Apostolova2013} or by an order of magnitude lower (\cite{Chhatkuli2013} reported 2.4~seconds per image).

\subsubsection{Evaluation on NLM Dataset}
\label{sec:evaluation-nlm}

Table~\ref{tab:eval-apostolova} shows the results of evaluating our proposed algorithm on the NLM CFS dataset (see Section~\ref{sec:datasets}) using the NLM evaluation procedure described in Section~\ref{sec:evaluation-methods}. We used the same parameter settings as in Section~\ref{sec:evaluation-imageclef} to demonstrate the generalization capability of our algorithm. We selected the most effective illustration classifiers using logistic regression and SVM, respectively. They both use \emph{simple11} features and the \emph{greedy} training set. For convenience, we also included the results reported in \cite{Apostolova2013} for a direct comparison with our approach.%
\footnote{The dataset reported in \cite{Apostolova2013} contains 400 images with 1764 ground-truth subfigures, so reported recall may be up to 0.4\% higher if evaluated on the 398 images of the dataset available to us.} 

\begin{table}
\caption{Evaluation results on the NLM CFS dataset \cite{Apostolova2013}. Precision (P), recall (R), and $F_1$ score are computed from the total number of ground-truth (G), detected (D), and true positive (T) subfigures.}
\label{tab:eval-apostolova}
\begin{center}
\begin{tabular}{lcccccc}
\hline
\textbf{Algorithm} & \textbf{G} & \textbf{D} & \textbf{T} & \textbf{P\%} & \textbf{R\%} & $\mathbf{F_1}$\textbf{\%}\\
\hline
Proposed (LogReg)                       & 1656 & 1550 & 1314 & 84.8 & \bf 79.4 & \textbf{82.0} \\
Proposed (SVM)                          & 1656 & 1584 & 1297 & 81.9 & 78.3 & 80.1 \\
Apostolova et al. \cite{Apostolova2013} & 1764 & 1482 & 1276 & \bf 86.1 & 72.3 & 78.6 \\
\hline
\end{tabular}
\end{center}
\end{table}

Results show that the relative performance of the proposed algorithm using different classifiers is consistent with evaluation results in Section~\ref{sec:evaluation-imageclef}. The proposed algorithm could detect 10\% more true positive subfigures than the image panel segmentation algorithm of Apostolova et al. \cite{Apostolova2013}, leading to a higher recall rate. On the other hand, precision is only slightly lower. Note that algorithm \cite{Apostolova2013} has been used as a component in NLM's CFS approach \cite{Santosh2015} referenced in Section~\ref{sec:evaluation-imageclef}.

\subsubsection{Illustration Classifier Accuracy}
\label{sec:evaluation-classifier}

To investigate the correlation of illustration classifier performance and effectiveness for CFS, we evaluated classification accuracy for the various classifier implementations considered in Section~\ref{sec:evaluation-imageclef} on the test dataset of the ImageCLEF 2015 multi-label image classification task \cite{Herrera2015}. Labels of test images were mapped to binary meta classes using the same procedure as described in Section~\ref{sec:illustration-classifier}, resulting in 497 images for \emph{first} and \emph{greedy} test sets, 428 images for \emph{majority}, and 398 images for \emph{unanimous} test set. Evaluation results are shown in Table~\ref{tab:classifier-results}. The decision threshold for logistic regression was set to 0.5 to provide a fair comparison with SVM. Internal parameters of SVM (box constraint $C$ and standard deviation $\sigma$ of RBF kernel) were optimized using two-fold cross-validation on the test set. 

\begin{table}
\caption{Classification accuracy on ImageCLEF 2015 multi-label image classification test dataset (497 images) for different implementation options of illustration classifier. Features and training sets are described in Section~\ref{sec:illustration-classifier}, LogReg = logistic regression.}
\label{tab:classifier-results}
\centering
\begin{tabular}{lllc}
\hline
\textbf{Classifier} & \textbf{Features} & \textbf{Training Set} & \textbf{Accuracy \%} \\
\hline
LogReg & simple2 & first         & 82.5 \\
LogReg & simple2 & majority      & 86.5 \\
LogReg & simple2 & unanimous     & \bf 88.2 \\
LogReg & simple2 & greedy        & 84.7 \\
\hline
LogReg & simple2 & greedy        & 84.7 \\
LogReg & simple11 & greedy       & 83.7 \\
SVM    & simple2 & greedy        & 84.3 \\
SVM    & simple11 & greedy       & 84.3 \\
SVM    & CEDD & greedy           & \bf 87.1 \\
SVM    & CEDD\_simple11 & greedy & 86.7 \\
\hline
\end{tabular}
\end{table}

The upper part of Table~\ref{tab:classifier-results} tells us that \emph{majority} and \emph{unanimous} training sets improve classification performance, although we know from Section~\ref{sec:evaluation-imageclef} that this does not help CFS effectiveness. From the lower part of Table~\ref{tab:classifier-results} we note that, interestingly, SVM does not perform better on \emph{simple2} features than logistic regression and causes only a modest improvement (around 3\%) on CEDD features (144-dimensional). This may indicate the need to select more discriminative features for this classification task in future work, although results of Section~\ref{sec:evaluation-imageclef} suggest that accuracy of the illustration classifier is not a critical factor of the proposed CFS algorithm.

\subsubsection{Limitations of CFS Algorithm}

\begin{figure}
\includegraphics[width=\textwidth]{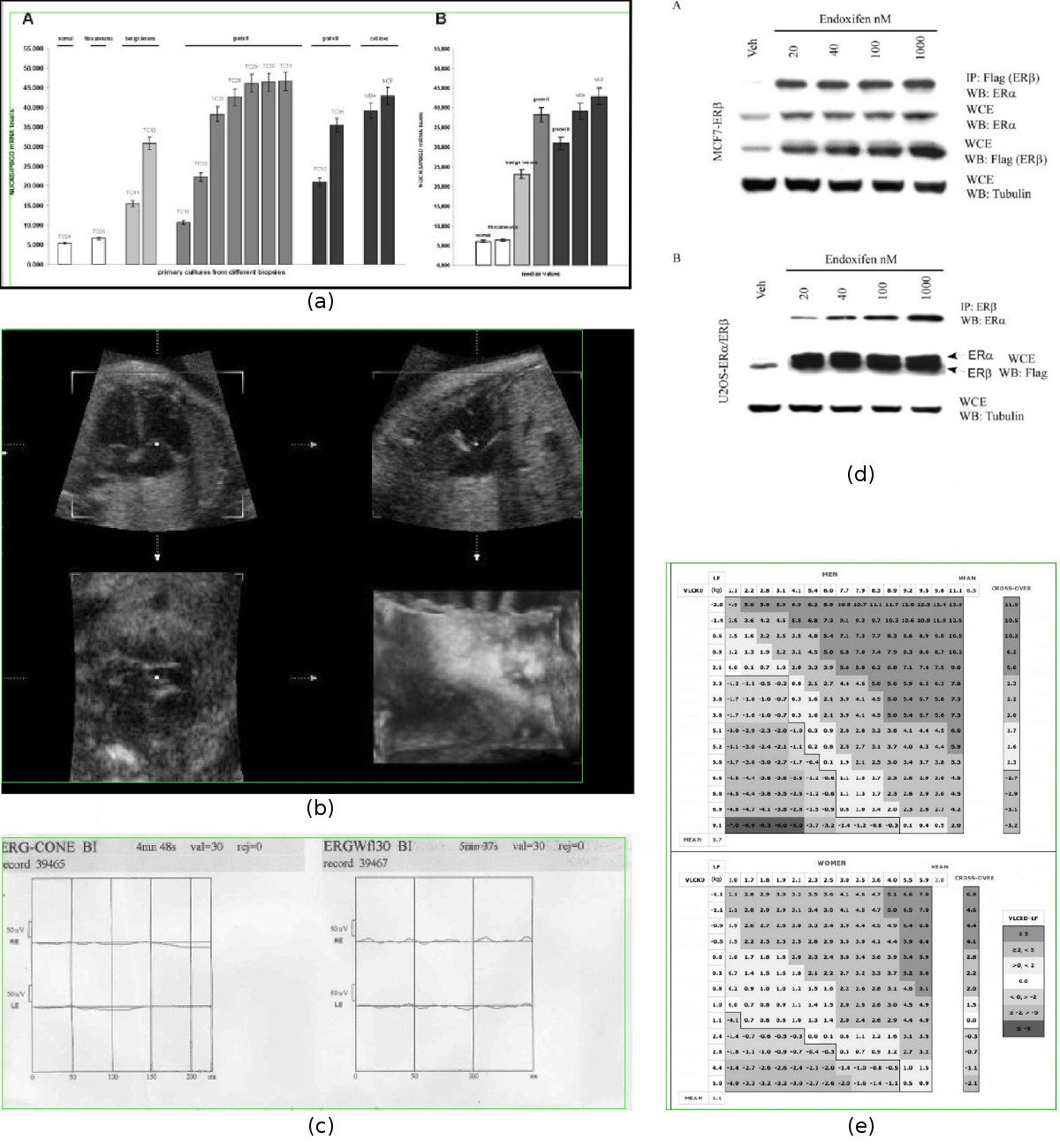}
\caption{Sample images of the compound figure separation test dataset \cite{Herrera2015} where our CFS algorithm failed: (a) dark figure border hinders separator band detection; (b) continuous horizontal or vertical separator edges do not exist; (c) separator band detection is hindered by noise after image binarization; (d) band-based over-segmentation results in many small subfigures that are discarded; (e) illustration without separator bands. Green lines indicate subfigure boundaries produced by our CFS algorithm.}
\label{fig:sample-failures}
\end{figure}

Figure~\ref{fig:sample-failures} shows some examples of test images where our algorithm did not perform as expected, for different reasons. In parts (a) and (e), the separator band detection algorithm fails due to the presence of dark lines, around the image and/or within the compound figure (which, ironically, was probably drawn to serve as visual separator between the subfigures). In part (b), the edge-based separation algorithm fails due to the lack of \emph{continuous} horizontal or vertical separator edges. 
In part (c), the presence of noise in the original image leads to an imperfect binarized version of the figure, which consequently impacts the performance of the separator band detection algorithm. Lastly, part (d) shows an example of over-segmentation, in which the proposed band-based algorithm first produces many subfigures, which  are subsequently discarded for being too small.

\subsubsection{Limitations of CFS Evaluation}

\begin{figure}
\begin{tabular}{cc}
\includegraphics[width=0.49\textwidth]{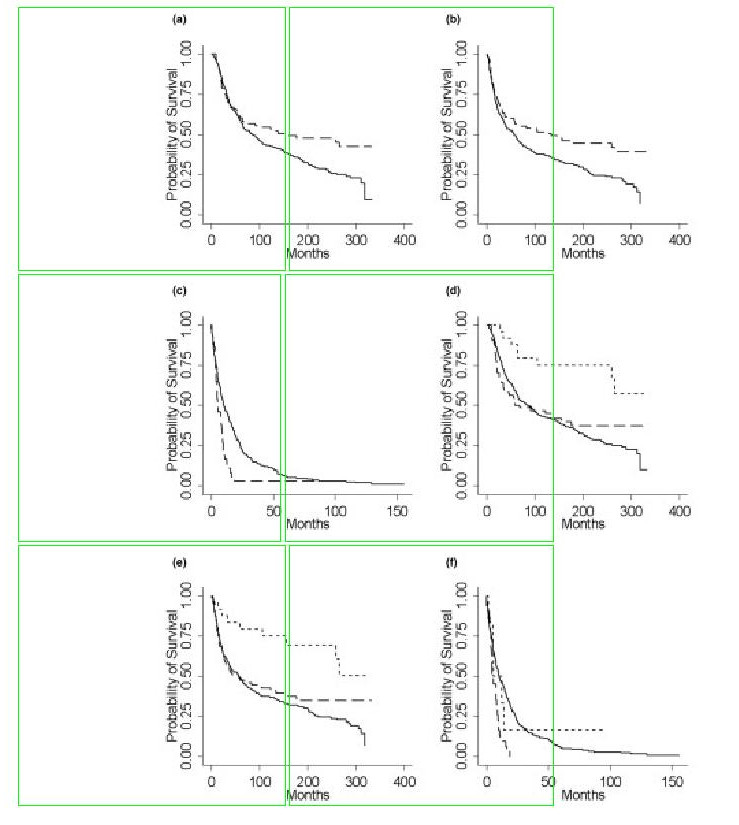} & \includegraphics[width=0.49\textwidth]{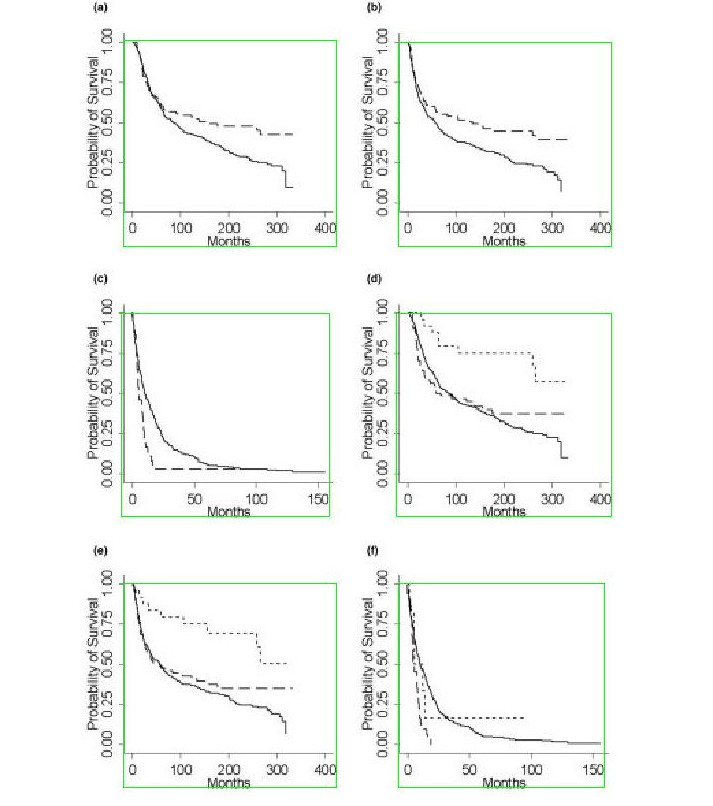}\\
(1) & (2) \\
\end{tabular}
\caption{(1) Example image of ImageCLEF CFS test dataset with imprecise ground-truth annotations. (2) Result  produced by our CFS algorithm, which was erroneously determined as having accuracy~0.}
\label{fig:wrong-cfs-annotation}
\end{figure}

The validity of a CFS evaluation procedure depends on both the quality of the test dataset, including ground-truth annotations, and the meaningfulness of the adopted performance metric. We recognized limitations in both aspects during experiments on the ImageCLEF CFS dataset. From the 260 images of the test set that received an accuracy of zero after being processed by our best CFS run (see Table~\ref{tab:evaluation-results}), we randomly selected 10 images and investigated the reason for failure. For three of them our CFS algorithm produced meaningful results, but errors in ground-truth annotations caused the ImageCLEF evaluation method to return accuracy~0, as illustrated for one of those images in Fig.~\ref{fig:wrong-cfs-annotation}.

\begin{figure}
\begin{tabular}{cc}
\includegraphics[width=0.49\textwidth]{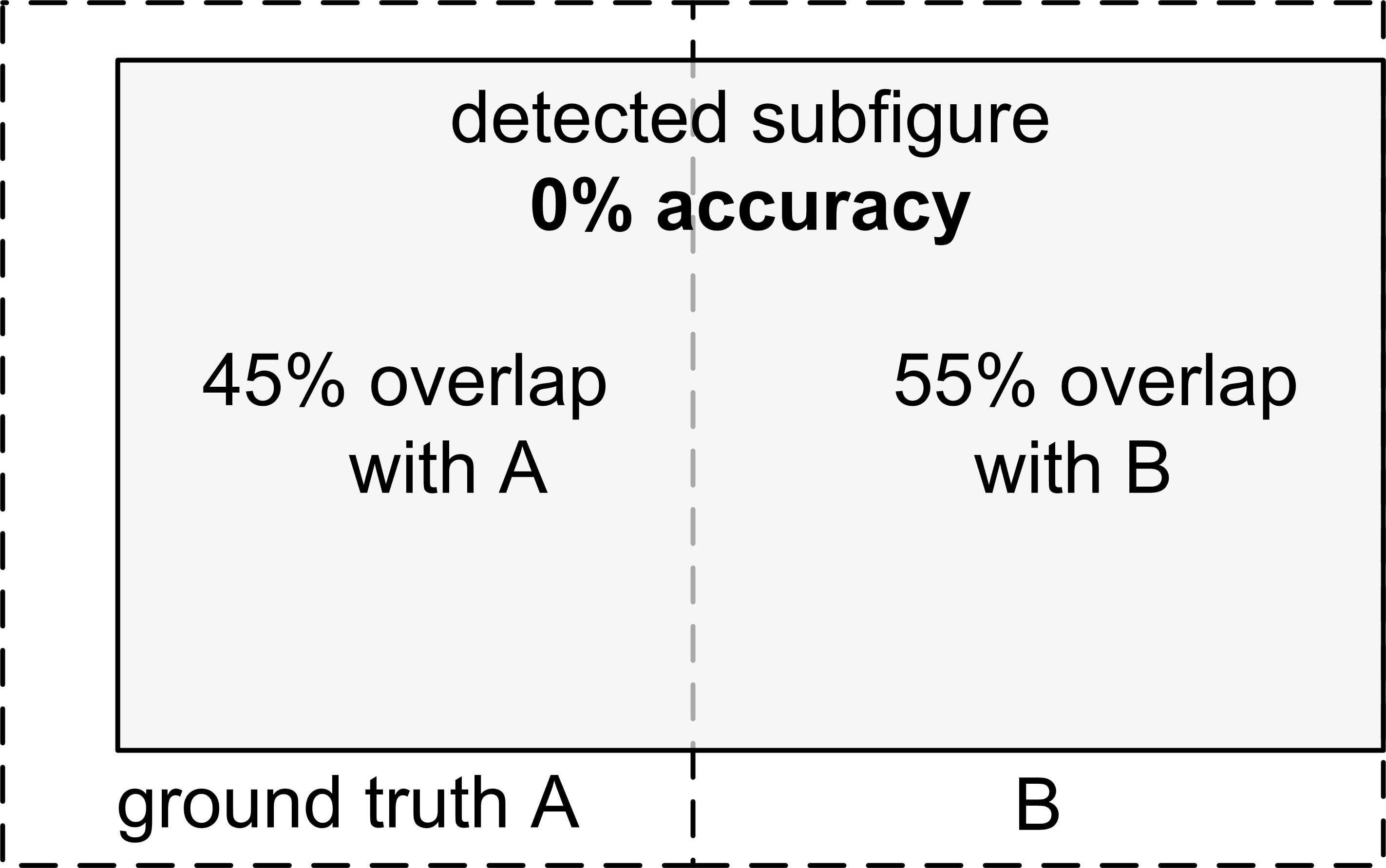} & \includegraphics[width=0.49\textwidth]{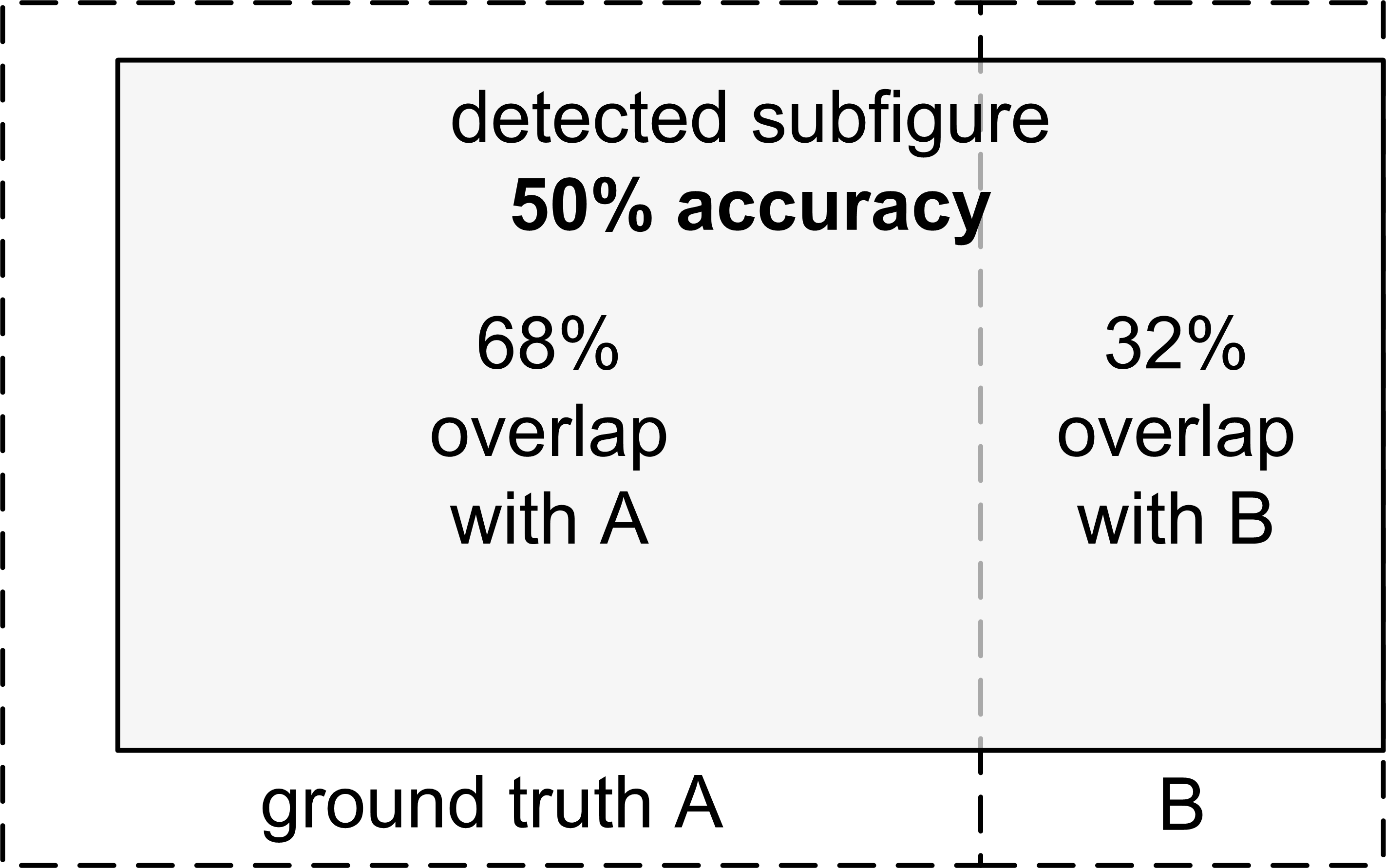}\\
(a) & (b) \\
\end{tabular}
\caption{Two similar under-segmentation cases lead to very different accuracy values according to ImageCLEF CFS evaluation procedure, because one of the ground-truth subfigures covers (a) less or (b) more than 2/3 of the single detected subfigure.}
\label{fig:cfs-accuracy-problem}
\end{figure}

In the second aspect, the ImageCLEF CFS evaluation procedure exhibits a notable instability with respect to under-segmentation: the CFS result in Fig.~\ref{fig:cfs-accuracy-problem}(a) is assigned an accuracy of zero, because none of the overlap ratios with the two ground-truth subfigures exceeds 2/3. On the other hand, for the similar situation in Fig.~\ref{fig:cfs-accuracy-problem}(b) the obtained accuracy is 50\%, because one of the ground-truth subfigures covers more than 2/3 of the single detected subfigure. Note that this problem does not occur with the NLM CFS evaluation procedure.

\subsection{CFC-CFS Chain}
\label{sec:cfc-cfs-experiments}

\begin{table}
\caption{Evaluation results of CFC-CFS chain for different algorithms and decision thresholds of the compound figure classifier (CFC). Decision thresholds are applicable to the logistic regression (LogReg) classifier only. The threshold marked by * was found to be optimal on the validation set. CR is the percentage of images classified as compound. In addition to accuracy on the total test set, accuracy values on the subsets of predicted compound (C) and non-compound (NC) images are shown.}
\label{tab:cfc-cfs-chain-results}
\centering
\begin{tabular}{lrr@{\hspace*{5mm}}rrr}
\hline
\multirow{2}{*}{\textbf{CFC}} & \multirow{2}{*}{\textbf{Threshold}} & \multirow{2}{*}{\textbf{CR\%}} & \multicolumn{3}{c}{\textbf{Accuracy\%}} \\
\cline{4-6}
& & & \textbf{C} & \textbf{NC} & \textbf{Total} \\
\hline
LogReg & 0.20  & 84 & 84.7 & 94.7 & 86.4\\ 
LogReg & *0.35 & 74 & 84.9 & 90.8 & 86.5\\ 
LogReg & 0.50  & 66 & 85.2 & 86.6 & 85.6\\ 
LogReg & 0.65  & 56 & 85.9 & 81.1 & 83.8\\ 
linear SVM & --  & 61 & 85.6 & 82.1 & 84.2\\ 
kernel SVM & --  & 74 & 84.4 & 95.6 & \textbf{87.3}\\ 
\hdashline
\emph{none} & 0 & 100 & 85.1 & -- & 85.1\\
\emph{ideal} & & 50 & 84.9 & 100  & 92.5\\ 
\hline
\end{tabular}
\end{table}

We used the CFC-CFS test dataset and evaluation procedure described in Sections~\ref{sec:datasets} and \ref{sec:chain-evaluation}, respectively, to evaluate the effectiveness of the proposed CFC-CFS process chain. Results obtained using the ImageCLEF CFS evaluation method are presented in Table~\ref{tab:cfc-cfs-chain-results}. For each of the three CFC algorithms (logistic regression, linear SVM, and kernel SVM) evaluated in Section~\ref{sec:cfc-experiments}, we applied the best-performing parameter settings according to Table~\ref{tab:cfc-results}. From these classifier algorithms, only logistic regression delivers predicted class probabilities, which allows to tune the effectiveness of the CFC-CFS chain by optimizing the decision threshold (Equation~\eqref{eq:decision-criterion} in Section~\ref{sec:cfc-cfs-chain}). Optimization was performed by evaluating CFC-CFS effectiveness on the CFC-CFS validation set for decision thresholds~$d$ in the range $0.2\le d\le 0.7$ using a step size of~0.05. The optimal value was found as $d=0.35$, corresponding to weight $\alpha=1.86$ of the misclassification loss matrix (Eq.~\eqref{eq:loss-matrix}). In Table~\ref{tab:cfc-cfs-chain-results}, we report results for four different decision thresholds on the test set. The optimal threshold selected during optimization on the validation set (indicated by *) also delivers best performance on the test set, confirming that improved performance for decision thresholds $d<0.5$ is not caused by overfitting the validation set.

Column CR (``compound rate'') of Table~\ref{tab:cfc-cfs-chain-results} shows the percentage of input images classified as \emph{compound} by the different CFC implementations. Separate accuracy values on the portions of the test set classified as compound and non-compound, respectively, indicate a natural trend: accuracy increases with decreasing size of the class-specific subset. For logistic regression, the increase of accuracy on the non-compound subset for shrinking decision thresholds overcompensates the moderate loss on the compound subset, improving total accuracy. As the decision threshold approaches zero, however, the number of predicted non-compound images and hence their effect on total accuracy becomes too small, leading to the observed local maximum of total accuracy for decision threshold $d=0.35$.

High CFC-CFS accuracy on the subset of predicted non-compound images can also be explained by a low false negative rate of CFC: false negatives are true compound images classified as non-compound, which are not sent through CFS processing and hence hurt effectiveness of the CFC-CFS chain. This explains the good performance of kernel SVM in Table~\ref{tab:cfc-cfs-chain-results}, although kernel SVM achieved inferior accuracy in CFC experiments (Table~\ref{tab:cfc-results}). From the three tested CFC algorithms, kernel SVM happened to have the lowest false negative rate at the cost of a high false positive rate, leading to a similar effect as decreasing the decision threshold for logistic regression.

From a wider perspective, however, effectiveness of CFC in the CFC-CFS process chain is rather limited when compared to processing all images of the test dataset with CFS only (indicated by classifier \emph{none} in Table~\ref{tab:cfc-cfs-chain-results}). In fact, our CFC implementations could improve CFC-CFS chain effectiveness by 2\% only, whereas an \emph{ideal} CFC algorithm that reproduces ground-truth class annotations would increase total accuracy by more than 7\%.

Finally we note that all pairwise differences of total accuracy values in Table~\ref{tab:cfc-cfs-chain-results}, which are mean values of accuracies determined for every input image, are statistically significant except for the difference between the first two lines in the table (logistic regression with decision thresholds 0.2 and 0.35, respectively). Significance has been tested at the 5\% significance level using a paired t-test.


\section{Conclusions}
\label{sec:conclusion}

In this paper we have proposed, implemented, tested, and evaluated a method to automatically classify and separate compound figures often found in scientific articles. The proposed method consists of two main steps: (i) a supervised compound figure classifier (CFC) discriminates between compound and non-compound figures using task-specific image features; and (ii) an image processing algorithm is applied to predicted compound images to perform compound figure separation (CFS). Combined, they are referred to as the \emph{CFC-CFS process chain}, to emphasize the dependencies and relationships between the two main blocks.

We have also introduced novel image features for compound figure classification and demonstrated that they can be used to achieve state-of-the-art CFC performance using well-known classifier algorithms.

Moreover, we have demonstrated that the proposed CFS algorithm outperforms state-of-the-art automatic and semi-automatic CFS approaches on two recently published biomedical datasets.

Lastly, we have established a method to evaluate the effectiveness of the CFC-CFS process chain and applied it to optimize the misclassification loss of CFC for maximal effectiveness in the process chain.

Future work might include algorithmic refinements to the CFS approach to address limitations (such as those illustrated in Figure~\ref{fig:sample-failures}), as well as implementation and testing of additional features and classification algorithms for CFC. When larger training datasets become publicly available, the use of deep learning techniques (e.g., convolutional neural networks) should also be considered.

%

\begin{acknowledgements}
We thank Sameer Antani (NLM) and the authors of \cite{Apostolova2013} for providing their compound figure separation dataset for evaluation purposes. We are grateful to Laszlo B{\"o}sz{\"o}rmenyi (ITEC, AAU) for valuable discussions and comments on this work.
\end{acknowledgements}


\bibliographystyle{spmpsci}
\bibliography{references}

%

\end{document}